\documentclass[10pt,journal]{IEEEtran}
\usepackage{amsmath}
\usepackage{amsfonts}
\usepackage{multirow}
\usepackage{booktabs}
\usepackage{amsthm}
\usepackage[noend]{algpseudocode}
\usepackage{algorithmicx,algorithm}
\usepackage{subfigure}
\usepackage{graphicx}
\usepackage{tabularx}
\usepackage{moresize}
\usepackage{xcolor}
\usepackage{tikz}
\usepackage[numbers,sort&compress]{natbib}
\usetikzlibrary{arrows,shapes,chains}

\ifCLASSINFOpdf
\else
\fi

\begin{document}
\title{Color Image and Multispectral Image Denoising Using Block Diagonal Representation}

\author{Zhaoming~Kong and Xiaowei~Yang ~\IEEEmembership{}
\IEEEcompsocitemizethanks{\IEEEcompsocthanksitem Zhaoming Kong (E-mail: kong.zm@mail.scut.edu.cn) and Xiaowei Yang (xwyang@scut.edu.cn) are with the school of Software Engineering, South China University of Technology, Guangdong Province, China.\protect\\}}


\maketitle

\begin{abstract}
Filtering images of more than one channel is challenging in terms of both efficiency and effectiveness. By grouping similar patches to utilize the self-similarity and sparse linear approximation of natural images, recent nonlocal and transform-domain methods have been widely used in color and multispectral image (MSI) denoising. Many related methods focus on the modeling of group level correlation to enhance sparsity, which often resorts to a recursive strategy with a large number of similar patches. The importance of the patch level representation is understated. In this paper, we mainly investigate the influence and potential of representation at patch level by considering a general formulation with block diagonal matrix. We further show that by training a proper global patch basis, along with a local principal component analysis transform in the grouping dimension, a simple transform-threshold-inverse method could produce very competitive results. Fast implementation is also developed to reduce computational complexity. Extensive experiments on both simulated and real datasets demonstrate its robustness, effectiveness and efficiency.
\end{abstract}


\IEEEpeerreviewmaketitle

\section{Introduction}
\IEEEPARstart
IMAGE denoising plays an important role in modern image processing systems. The past few decades witness great achievements in this field \cite{Milanfar2011A}, and methods that utilize self-similarity and non-local characteristics of natural images have drawn much attention due to their simplicity and effectiveness. Recently, great achievement is made by the well-known BM3D algorithm \cite{Dabov2007Image} which combines the nonlocal filters \cite{Buades2005A} and transform domain techniques \cite{Yaroslavsky2001Transform}. Methods that share similar idea and procedure of BM3D are widely adopted to handle grayscale image \cite{Dabov2010Video, gu2014weighted, rajwade2013image}. When the input is color image (sRGB) or multispectral image that contains rich information and delivers more faithful representation for real scenes, directly applying the grayscale denoising algorithm to each channel \cite{Buades2005A} or spectral band \cite{yan2013nonlocal} often fails to produce satisfactory results, and therefore efforts to understand and address noise reduction issue have been made from several different perspectives.\\
\indent First, two main solutions are proposed to improve the channel-by-channel or band-by-band approach. The first strategy proposes to transform the original image into a less correlated color or band space, such that denoising in each transformed channel or band could be performed independently. the representative work is the color BM3D (CBM3D) which applies BM3D to the luminance-chrominance space \cite{dabov2007color}. The second strategy jointly characterizes the RGB channels or bands for better use of spectral correlation. Methods that fall into such category are widely considered with different priors and regularization. Briefly, \cite{dai2013multichannel} proposes a multichannel nonlocal fusion (MNLF) approach, \cite{tu2014collaborative} introduces color line to model the correlation among neighbouring pixels and channels, and \cite{zhong2013multiple} considers the spatial and spectral dependencies. Sparse and low rank priors are also adopted in several competitive methods \cite{xu2017multi,he2016total,TrilateralXu}. Besides, to avoid vectorization of image patch, some recent works incorporate tensor representation \cite{Kolda2009Tensor} with higher order singular value decomposition (HOSVD) \cite{Tucker1966Some, Lathauwer2000A, rajwade2013image}, low rank tensor approximation (LRTA) framework \cite{renard2008denoising}, Laplacian Scale Mixture modeling \cite{dong2015low} and Hyper-Laplacian regularization \cite{chang2017hyper}.\\
\indent In addition to the design of denoising strategy, noise modeling is also important. Most of existing methods consider additive white Gaussian noise (AWGN) and some efficient noise estimation methods \cite{chen2015efficient, Xinhao2013Single} can be employed. Besides, some non i.i.d. Gaussian denoisers are proposed for filtering Poission noise \cite{Zhang2008Wavelets},
 mixed Gaussian and impulsive noise \cite{Xu2016patch} and stripe noise \cite{chang2015anisotropic}. In fact, noise in real-world images may be multiplicative and signal dependent \cite{ramanath2005color}, making noise modeling and estimation much more complex and challenging. \cite{nam2016holistic} and \cite{wong2016turbo} consider the non-linear processing steps in the camera pipeline in the noise model, and \cite{mosseri2013combining, Luo2015Adaptive, xu2018external} combine external and internal priors. Some methods \cite{Lebrun2015Multiscale, Lebrun2015The, chen2018denoising, zhu2016noise}, including the well-known software toolbox Neat Image (NI)\footnote{Neatlab ABSoft. https://ni.neatvideo.com/home} are developed for noise reduction of real-world images. Apart from the conventional transform-domain approaches, many recent competitive methods \cite{burger2012image, zhang2017beyond, chang2018hsi, Zhang2018FFDNet} are based on the advent of deep learning technique as a powerful feature extraction tool.\\
\indent In order to compare different denoising methods, several real-world color image and multispectral image datasets \cite{Anaya2014RENOIR,nam2016holistic,plotz2017benchmarking,DatasetXu,abdelhamed2018high,chakrabarti2011statistics,KongData} of various scenes are constructed, and each scene of a color image includes noisy and "ground-truth" image pairs. A simple and reasonable approach adopted by \cite{Anaya2014RENOIR,nam2016holistic,DatasetXu,KongData} to obtain "ground-truth" image is to capture the same and unchanged scene for many times and compute their mean image. Different from the image averaging approach, \cite{plotz2017benchmarking} utilizes the Tobit regression to estimate the parameters of the noise process by accessing only two images, and \cite{abdelhamed2018high} generates a high quality smartphone image denoising dataset with careful post-processing. Interestingly, recent experiments on real-world datasets show that BM3D based methods \cite{dabov2007color, Maggioni2013Nonlocal} still demonstrate the most competitive performance in terms of both effectiveness and efficiency. The implementation details are unknown and there is debate \cite{chatterjee2010denoising,levin2011natural} that they may have touched the ceiling of image denoising.\\
\indent Many competitive methods \cite{xu2017multi,chang2017hyper,TrilateralXu} attempt to approach the optimal performance by modeling the redundancy and correlation at group level with some iterative strategies \cite{romano2017little} and a large number of similar patches. However, influence of the patch level representation is less carefully studied. Although the use of tensor representation may help preserve some structure information, the straightforward folding and unfolding operation may not fully exploit the relationship among all channels or spectral bands. In this paper, we investigate the potential and influence of patch level representation, and establish a general formulation with block diagonal matrix. We demonstrate that the combination of a proper global patch basis and local PCA can produce very competitive performance in terms of both efficiency and effectiveness. Extension to non-Gaussian noise in hyperspectral image is also discussed. Efforts have been made to reduce computational complexity, and all results reported in our paper could be reproduced very efficiently.\\
\indent The rest of this paper is organized as follows: In Section II, related work is studied. In Section III, general formulation and the proposed denoising method are introduced. Section IV presents experiments on both simulated and real-world datasets. Conclusions are drawn in Section V.

\section{Related Works and Formulation}
\subsection{Notations}
\indent Tensor is a multidimensional array, also known as a multi-way array, and its \textsl{order} is defined as the number of its dimension. In this paper, we mainly adopt the mathematical notations and preliminaries of tensors from \cite{Kolda2009Tensor}. Vectors and matrices are first- and second- order tensors which are denoted by boldface lowercase letters $\mathbf{a}$ and capital letters $\mathbf{A}$, respectively. A higher order tensor (the tensor of order three or above) is denoted by calligraphic letters, e.g., $\mathcal{A}$. An $N$th-order tensor is denoted as $\mathcal{A} \in \mathbb{R}^{I_1\times I_2\times\cdots\times I_N}$. The $n$-mode product of a tensor $\mathcal{A}$ by a matrix $\mathbf{U}\in \mathrm{R}^{P_n\times I_n}$, denoted by $\mathcal{A}\times _n\mathbf{U}$ is also a tensor. The mode-$n$ matricization or unfolding of $\mathcal{A}$, denoted by $\mathbf{A}_{(n)}$, maps tensor elements $(i_1,i_2,\ldots,i_N)$ to matrix element $(i_n,j)$ where $j=1+\sum_{k=1,k\neq n}^{N}(i_k-1)J_k$, with $J_k = \prod_{m=1,m\neq n}^{k-1}I_m$. The Frobenius norm of a tensor $\mathcal{A} \in \mathbb{R}^{I_1\times I_2\times\cdots\times I_N}$ is defined as $\|\mathcal{A}\|_F = \sqrt{\sum_{i_1=1}...\sum_{i_N=1}\mathcal{A}_{i_1...i_N}^2}$.
\subsection{Framework and Problem Formulation}
\begin{table}[htbp]
\ssmall
  \centering
    \caption{Techniques and priors adopted by different methods based on the grouping-collaborative filtering-aggregation framework. LR: Low Rank; SC: Sparse Coding; HT: Hard-thresholding; WF: Wiener Filter \cite{Dabov2007Image}}
    \begin{tabular}{ccccccccc}
    \toprule
    Method & LLRT \cite{chang2017hyper}  & WTR1 \cite{Li2018Weighted}   & TWSC \cite{TrilateralXu}  & NCSR \cite{Dong2013Sparse} & CBM3D & Proposed\\
    \midrule
    Techique & LR   &  LR     & SC    & SC & HT+WF & HT\\
    \bottomrule
    \end{tabular}%
  \label{Table_technique_briefing}%
\end{table}%

\begin{figure}[htbp]
\graphicspath{{Illustration/Other/}}
\centering
\subfigure{
\label{Fig4}
\includegraphics[width=3.6in]{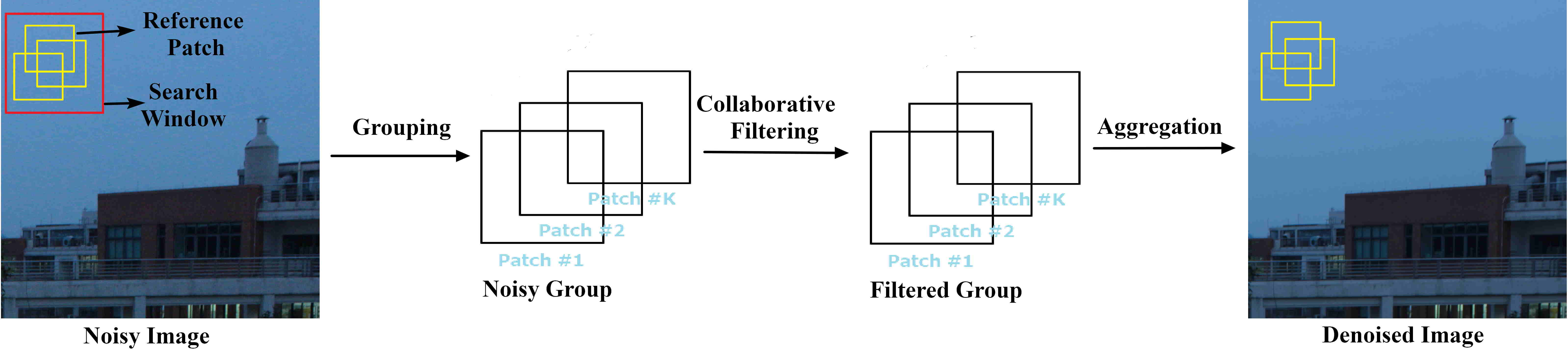}}

\caption{Flowchart of grouping-collaborative filtering-aggregation framework.}

\label{Fig_illus_paradigm}
\end{figure}

The most popular and successful framework credited to \cite{Dabov2007Image} follows three consecutive steps: grouping, collaborative filtering and averaging. The flowchart of this paradigm is illustrated as Fig. \ref{Fig_illus_paradigm}. Specifically, given a reference patch $\mathcal{P}_{ref}$, the grouping step stacks some similar overlapping patches located in a local window $\Omega_{SR}$ into a group represented by matrix $\mathbf{G}$ or higher order tensor $\mathcal{G}$ with certain matching criteria \cite{Foi2007Pointwise, Buades2016Patch, Foi2016Foveated}. One simple and commonly adopted metric is Euclidean distance measured by $\|\mathcal{P}_{ref} - \mathcal{P}\|_F^2$, $\forall \mathcal{P} \in \Omega_{SR}$. Collaborative filtering is then performed on group $\mathbf{G}$ to utilize the nonlocal similarity feature and estimate clean underlying patches from noisy observation, and it can be generally formulated as
\begin{equation}\label{collarborative_filtering}
  \hat{\mathbf{G}} = \mathop{\arg\min_{\mathbf{G}_c}} \| \mathbf{G}_n - \mathbf{G}_c \|_{F}^2 + \rho\cdot\Psi(\mathbf{G}_c)
\end{equation}
where $\mathbf{G}_n$ and $\mathbf{G}_c$ are noisy and underlying clean group of patches, respectively, $\| \mathbf{G}_n - \mathbf{G}_c \|_{F}^2$ measures the conformity between $\mathbf{G}_c$ and $\mathbf{G}_n$, and $\Psi(\mathbf{G}_c)$ represents certain priors \cite{Ling2017From}. Low rank approximation is adopted in \cite{Dong2013Nonlocal, xu2017multi, chang2017hyper, Li2018Weighted} based on nuclear norm minimization \cite{Cai2008A} with $\Psi(\mathbf{G}_c) =  \|\mathbf{G}_c\|_{\ast}$, or tensor trace norm \cite{Liu2013Tensor} with $\Psi(\mathcal{G}_c) =  \sum_{n=1}^{N} \alpha_n \|\mathbf{G_c}_{(n)}\|_{\ast}$. Authors in \cite{Elad2006Image, TrilateralXu, Dong2013Sparse, Mairal2009Sparse} utilize sparse coding scheme that represents $\hat{\mathbf{G}}$ with a dictionary $\mathbf{D}$ and sparse coding atoms $\mathbf{C}$ by minimizing
\begin{equation}\label{sparse_coding}
  \hat{\mathbf{C}} = \mathop{\arg\min_{C}} \|\mathbf{G}_n - \mathbf{D}\mathbf{C}\|_F^2 + \lambda \|\mathbf{C}\|_1
\end{equation}
 The state-of-the-art BM3D and HOSVD algorithms attempt to model sparsity in the transform domain by shrinking coefficients $\mathcal{T}(\mathcal{G}_n) $ under a pre-defined threshold $\tau$ via
 \begin{equation}\label{hard_thresholding}
  \mathcal{T}(\mathcal{G}_{ht})=\left\{
\begin{aligned}
\mathcal{T}(\mathcal{G}_n), \quad |\mathcal{T}(\mathcal{G}_n)| \geq \tau \\
0, \quad |\mathcal{T}(\mathcal{G}_n)| < \tau
\end{aligned}
\right.
 \end{equation}
 Some representative techniques and priors are listed in Table \ref{Table_technique_briefing}. After collaborative filtering, the estimated clean patches are averagely written back to their original location to further smooth out noise. More specifically, every pixel $\hat{p}_i$ of the denoised image is the (weighted) average of all pixels at the same position of filtered group $\hat{\mathbf{G}}$, which can be formulated as
 \begin{equation}\label{aggregation}
   \hat{p}_i = \sum_{\hat{p}_{i_k} \in \hat{\mathbf{G}}} w_{i_k} \hat{p}_{i_k}
 \end{equation}
 where $w_{i_k}$ and $\hat{p}_{i_k}$ denote weight and pixel, respectively.

\subsection{Multiway filtering technique}
In this paper, we focus on the collaborative filtering step with transform domain technique, and regardless of different categorization criteria, it can be roughly modeled as an inverse problem with multiway filtering technique \cite{muti2008lower}, which applies a specific transform and constraint to each dimension of a group of similar image patches. The organization of image patches and choice of transform and constraint often determine the denoising strategy, and the variations are quite extensive \cite{dabov2007color,rajwade2013image,dong2015low, Li2018Weighted}. In this subsection we introduce the representative 4DHOSVD and CBM3D for color image denoising, and our analysis can be extended to multispectral image with slight modification. For simplicity, a group of similar patches $\mathcal{P}$ is denoted by $\mathcal{G}$.\\
\indent Both CBM3D and 4DHOSVD can be represented with the fourth-order tensor decomposition framework in \cite{Kolda2009Tensor} via
\begin{equation}\label{4dhosvd}
  \mathcal{C} = \mathcal{G} \times _1\mathbf{U}_{row}^T \times _2\mathbf{U}_{col}^T \times _3\mathbf{U}_{color}^T \times _4\mathbf{U}_{group}^T
\end{equation}
where $\mathcal{C}$ is core tensor (coefficient), $\mathbf{U}_{row}$, $\mathbf{U}_{col}$, $\mathbf{U}_{color}$, and $\mathbf{U}_{group}$ are corresponding mode transforms. The major difference between CBM3D and 4DHOSVD is that CBM3D uses pre-defined discrete cosine transform (DCT) and opponent color space represented as a $3\times3$ matrix
\begin{equation}\label{opp_color_mode}
  \mathbf{U}_{color}^T = \left(
    \begin{array}{ccc}
      \textbf{1/3} & \textbf{1/3} & \textbf{1/3} \\
      0.5 & 0 & -0.5 \\
      0.25 & -0.5 & 0.25 \\
    \end{array}
  \right)
\end{equation}
specifically, the first slice of each patch in the new color space can be regarded as luminance channel, and the other two slices as chrominance channel. CBM3D is very efficient because it does not have to train local transforms, and its grouping step is performed only on the luminance channel. For 4DHOSVD, all mode transform matrices including $\mathbf{U}_{color}$ are learned by solving
\begin{equation}\label{optimize_4dhosvd}
\begin{split}
   &  \min \|\mathcal{G} - \mathcal{C} \times _1\mathbf{U}_{row} \times _2\mathbf{U}_{col} \times _3\mathbf{U}_{color} \times _4\mathbf{U}_{group}\|_F^2\\
    & s.t  \qquad \mathbf{U}_{row}^T\mathbf{U}_{row} = \mathbf{I},  \qquad \mathbf{U}_{col}^T\mathbf{U}_{col} = \mathbf{I} \\
    & \quad\qquad \mathbf{U}_{color}^T\mathbf{U}_{color} = \mathbf{I}, \qquad \mathbf{U}_{group}^T\mathbf{U}_{group} = \mathbf{I}
\end{split}
\end{equation}
Compared with CBM3D, 4DHOSVD imposes a stronger constraint and requires the orthogonality of color space transform.
\section{Image Denoising Using Block Diagonal Representation}
\indent Using a 4D transform in equation (\ref{4dhosvd}) for CBM3D may be a little confusing, because after a certain color space transform, the original R, G, B channels are computed separatedly in the new color space, which also holds for 4DHOSVD if all mode transforms are obtained. Therefore, it may be expected to re-formulate (\ref{4dhosvd}) as independent channel-wise (third-order tensor) transform. In this section, we first generalize patch-level representation via block diagonal matrix, then discuss the proper choice for patch-level basis, and explain how it could be properly incorporated into the block diagonal representation and efficiently applied to image denoising.
\subsection{Block Diagonal Representation and Formulation of f-diagonal Tensor Decomposition}
\noindent We notice that (\ref{4dhosvd}) can be rewritten as
\begin{equation}\label{4dhosvd_reform}
  \mathcal{C} = (\mathcal{G} \times _3\mathbf{U}_{color}^T) \times _1\mathbf{U}_{row}^T \times _2\mathbf{U}_{col}^T \times _4\mathbf{U}_{group}^T
\end{equation}
where $\mathcal{G} \times _3\mathbf{U}_{color}^T$ is equivalent to applying $\mathbf{U}_{color}$ to each patch of $\mathcal{G}$ via
\begin{equation}\label{4dhosvd_patch}
  (\mathcal{G} \times _3\mathbf{U}_{color}^T)_i = \mathcal{P}_i \times _3\mathbf{U}_{color}^T, \; i = 1,2,...,K
\end{equation}
where $K$ and $\mathcal{P}_i$ represent the number of similar patches and the $i$-th patch of $\mathcal{G}$, respectively. For simplicity, we use $\hat{\mathcal{P}}_i$ to denote $\mathcal{P}_i \times _3\mathbf{U}_{color}^T$. Then we could define the block diagonal operator $bdiag(\hat{\mathcal{P}}_i)$ via
\begin{equation}\label{block_diag_matrix}
  bdiag(\hat{\mathcal{P}}_i) = \left(
                       \begin{array}{ccc}
                         \hat{\mathcal{P}}_i(:,:,1) &  &  \\
                          & \hat{\mathcal{P}}_i(:,:,2) &  \\
                          &  &  \hat{\mathcal{P}}_i(:,:,3)\\
                       \end{array}
                     \right)
\end{equation}
where each matrix on the diagonal position is a linear combination of all frontal slices of $\mathcal{P}_i$ via
\begin{equation}\label{linear_p}
  \hat{\mathcal{P}}_i(:,:,k) = \sum_{j = 1}^3 \mathbf{U}_{color}(j,k) \mathcal{P}_{i}(:,:,j), \; k = 1, 2, 3
\end{equation}
according to (\ref{4dhosvd_patch}) and (\ref{block_diag_matrix}), $\mathcal{G} \times _3\mathbf{U}_{color}^T$ can be denoted as an f-diagonal tensor $fdiag(\hat{\mathcal{G}})$
\begin{equation}\label{block_diag_tensor}
  fdiag(\hat{\mathcal{G}}) = \left(
                       \begin{array}{ccc}
                         \hat{\mathcal{G}}_i(:,:,1,:) &  &  \\
                          & \hat{\mathcal{G}}_i(:,:,2,:) &  \\
                          &  &  \hat{\mathcal{G}}_i(:,:,3,:)\\
                       \end{array}
                     \right)
\end{equation}
where $\hat{\mathcal{G}} = \mathcal{G} \times _3\mathbf{U}_{color}^T$. Based on (\ref{4dhosvd_reform}) and (\ref{block_diag_tensor}), the 4D transform (\ref{4dhosvd}) is equivalent to
\begin{equation}\label{block_diag_tensor_decomp}
\begin{split}
   & \mathcal{C} = fdiag(\hat{\mathcal{G}}) \times _1bdiag(\mathbf{U}_{row}^T) \times _2bdiag(\mathbf{U}_{col}^T)\times _3\mathbf{U}_{group}^T
\end{split}
\end{equation}
where we make an abuse use of (\ref{block_diag_matrix}) to denote $bdiag(\mathbf{U}_{row})$ and $bdiag(\mathbf{U}_{col})$ as
\begin{equation}\label{block_diag_U_row}
  bdiag(\mathbf{U}) = \left(
                       \begin{array}{ccc}
                          \mathbf{U} &  &  \\
                          & \mathbf{U} &  \\
                          &  &  \mathbf{U} \\
                       \end{array}
                     \right)
\end{equation}
The same group representation $\mathbf{U}_{group}$ is applied for all frontal slices in (\ref{block_diag_tensor}) mainly because of two reasons. First, the patch-wise similarity often used in the grouping process does not guarantee slice-wise simialrity. Furthermore, according to equation (\ref{linear_p}), the new color space are a linear combination of R, G, B channels, thus are not totally de-correlated. Therefore, there exists a trade-off between the slice-wise and patch-wise relationship. Interestingly, CBM3D subtly takes care of this issue by considering only the luminance channel similarity. To narrow such gap, a suitable alternative to utilizing more group-level information (grouping more patches), is the recursive use of patch-level correlation via block circulant representation (BCR) \cite{Mazancourt1983The}.\\
\indent Specifically, for each patch $\mathcal{P}_i \in \mathbb{R}^{ps \times ps \times 3}$ of color image, its BCR $bcirc(\mathcal{P}_i)$ is a block circulant matrix \cite{Mazancourt1983The} of size $3ps \times 3ps$ defined by
\begin{equation}\label{bcirc_matrix}
  bcirc(\mathcal{P}_i) = \left(
                           \begin{array}{ccc}
                             \mathbf{P}_i(:,:,1) & \mathbf{P}_i(:,:,3) & \mathbf{P}_i(:,:,2) \\
                             \mathbf{P}_i(:,:,2) & \mathbf{P}_i(:,:,1) & \mathbf{P}_i(:,:,3) \\
                             \mathbf{P}_i(:,:,3) & \mathbf{P}_i(:,:,2) & \mathbf{P}_i(:,:,1) \\
                           \end{array}
                         \right)
\end{equation}
thus, similar to (\ref{block_diag_tensor}), a block circulant tensor $bcirc(\mathcal{G})$ of size $3ps\times 3ps \times K$ can be denoted as
\begin{equation}\label{bcirc_tensor}
  bcirc(\mathcal{G}) = \left(
                           \begin{array}{ccc}
                             \mathcal{G}(:,:,1,:) & \mathcal{G}(:,:,3,:) & \mathcal{G}(:,:,2,:) \\
                             \mathcal{G}(:,:,2,:) & \mathcal{G}(:,:,1,:) & \mathcal{G}(:,:,3,:) \\
                             \mathcal{G}(:,:,3,:) & \mathcal{G}(:,:,2,:) & \mathcal{G}(:,:,1,:) \\
                           \end{array}
                         \right)
\end{equation}
\begin{figure}[htbp]
\graphicspath{{Illustration/Other/}}
\centering
  \includegraphics[width=3.38in]{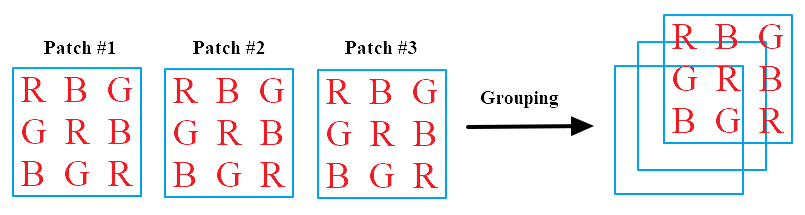}
  \caption{Block circulant representation of a group of similar patches.}

  \label{Fig_illus_bcirc_group}
\end{figure}

Fig. \ref{Fig_illus_bcirc_group} gives a straightforward illustration of (\ref{bcirc_matrix}) and (\ref{bcirc_tensor}), and some interesting feature is given in Appendix \ref{appendix_a}.
\newtheorem{myThm}{Theorem}[]
Following the idea of multiway transform technique in (\ref{4dhosvd}) and based on (\ref{group_equal}), the third-order tensor decomposition of (\ref{bcirc_tensor}) is
\begin{equation}\label{hosvd_bcirc_tensor}
  \mathcal{C}_{bcirc} = bcirc(\mathcal{G}) \times _1\mathbf{U}_{bcirc_{row}}^T \times _2\mathbf{U}_{bcirc_{col}}^T \times _3\mathbf{U}_{group}^T
\end{equation}
Obtaining two factor matrices $\mathbf{U}_{bcirc_{row}}$ and $\mathbf{U}_{bcirc_{col}}$ requires the eigenvalue decomposition of two large block circulant matrices. Using fast Fourier transform (FFT) \cite{rojo2004some}, the block circulant tensor decomposition problem in (\ref{hosvd_bcirc_tensor}) can be re-formulated as following f-diagonal tensor decomposition in the Fourier domain\footnote{More details are given in Appendix \ref{appendix_b}.}
\begin{equation}\label{hosvd_fdiag_tensor}
  \mathcal{C}_{fdiag} = fdiag(\hat{\mathcal{G}}) \times _1\mathbf{U}_{fdiag_{row}}^T \times _2\mathbf{U}_{fdiag_{col}}^T \times _3\mathbf{U}_{group}^T
\end{equation}
Where $\hat{\mathcal{G}} = \mathcal{G}\times _3\mathbf{W}$, and $\mathbf{W}$ is the FFT matrix defined as
 \begin{equation}\label{fft_matrix}
   \left(
     \begin{array}{ccc}
       \textbf{1} & \textbf{1} & \textbf{1} \\
       1 & -0.5 - 0.8660i & -0.5 + 0.8660i \\
       1 & -0.5 + 0.8660i & -0.5 - 0.8660i \\
     \end{array}
   \right)
 \end{equation}
The conjugate feature of $\mathbf{W}$ indicates that whichever the threshold technique (low-rank or hard-threshold) is adopted, only the first and second slices in the Fourier domain need to be computed. This strategy can be extended to handle multispectral images by considering only the first $\lfloor N/2 \rfloor + 1$ slices in the Fourier domain, where $N$ is the number of spectral bands.
\subsection{Equivalence of f-diagonal Tensor Decomposition in the Fourier Domain}
\indent In fact, the computation of (\ref{hosvd_fdiag_tensor}) does not require explicit formulation of f-diagonal tensor in the Fourier domain. We introduce t-SVD or tensor-SVD \cite{kilmer2011factorization, Kilmer2013Third}, a new third-order tensor decomposition framework that demonstrates competitive performance in many applications \cite{hao2013facial, zhang2016exact, zhou2017outlier}. t-SVD mainly relies on the definition of t-product $*$ between two third-order tensors using (\ref{block_diag_matrix}) in the Fourier domain.

\newtheorem{myDef}{Definition}[]

\begin{myDef}[t-product]
\label{t-product}
The t-product $*$ between two third-order tensor  $\mathcal{A} \in \mathbb{R}^{n_1\times n_2 \times n_3}$ and $\mathcal{B} \in \mathbb{R}^{n_1\times n_4 \times n_3}$ is an $n_1\times n_4 \times n_3$ tensor $\mathcal{C} = \mathcal{A} * \mathcal{B}$ given by
\begin{equation}\label{bdiag_t-product}
  bdiag(\hat{\mathcal{C}}) = bdiag(\hat{\mathcal{A}}) bdiag(\hat{\mathcal{B}})
\end{equation}
where $\hat{\mathcal{A}} = \mathcal{A} \times _3\mathbf{W}$ and $\hat{\mathcal{B}} = \mathcal{B} \times _3\mathbf{W}$.
\end{myDef}
\noindent The t-SVD of a third-order tensor can be defined as the t-product of three third-order tensor via
\begin{equation}\label{t-SVD}
  \mathcal{A} = \mathcal{U} * \mathcal{S} * \mathcal{V}^T
\end{equation}
which can be computed as
\begin{equation}\label{bdiag_t-SVD}
  bdiag(\hat{\mathcal{A}}) = bdiag(\hat{\mathcal{U}})bdiag(\hat{\mathcal{S}})bdiag(\hat{\mathcal{V}})^T
\end{equation}
with an abuse use of t-product, the core tensor can be obtained by
\begin{equation}\label{compute_C_4D}
  \mathcal{C} = \mathcal{G} * _1\mathcal{U}_{row}^T * _2\mathcal{U}_{column}^T * _3\mathcal{U}_{group}^T
\end{equation}
where $\mathcal{G} * _i\mathcal{U}^T$ can be computed as the $i$-th mode tensor-matrix product in the Fourier domain. $\mathcal{U}_{group}$ is a third-order tensor with its first frontal slice equal to $\mathbf{U}_{group}$ and other frontal slices equal to $\mathbf{0}$. Furthermore, $\mathcal{U}_{row}$ and $\mathcal{U}_{column}$ can be obtained by minimizing a non-local t-SVD (NL-tSVD) problem
\begin{equation}\label{nonlocal_tSVD}
\begin{split}
  & \min\sum_{i=1}^{K}\|\mathcal{P}_i - \mathcal{U}_{row}*\mathcal{S}_{i}*\mathcal{U}_{column}^T\|\\
  & s.t \quad \mathcal{U}_{row}^T * \mathcal{U}_{row} = \mathcal{I}, \quad \mathcal{U}_{column}^T * \mathcal{U}_{column} = \mathcal{I}
\end{split}
\end{equation}
After some threshold technique, the filtered group $\mathcal{G}_{filtered}$ can be obtained by the inverse transform of (\ref{compute_C_4D}) via
\begin{equation}\label{compute_filtered_group}
  \mathcal{G}_{filtered} = \mathcal{C} * _1\mathcal{U}_{row} * _2\mathcal{U}_{column} * _3\mathcal{U}_{group}
\end{equation}
\subsection{Threshold technique}
For multiway filtering approaches described in ($\ref{4dhosvd}$), there are roughly two strategies to encourage the sparsity of linear approximation in the transform domain: threshold core tensor $\mathcal{C}$ (via L0-norm \cite{qi2018multi}, Wiener filter \cite{letexier2008noise,rajwade2013image}, soft- or hard-thresholding \cite{rajwade2013image}), and threshold factor matrices via low rank prior \cite{fu20163d}. Directly modeling each mode of 4D data with low rank prior raises two major concerns. First, the possible combination of rank estimation along each mode is extensive, and it risks falling into the unbalance trap introduced in \cite{bengua2017efficient}. Briefly, given a 4D group of $K$ patches $\mathcal{G}\in \mathbb{R}^{8\times 8 \times N \times K}$, its first mode unfolding is a skinny matrix $\mathbf{G}_1 \in \mathbb{R}^{8\times (8NK)}$, since $8NK\gg8$, then the rank of $\mathcal{G}$ along the first mode is assumed very low, which risks the loss of more information \cite{bengua2017efficient}. To solve these issues, although not explicitly stated, some methods \cite{dong2015low, chang2017hyper} reshape $\mathcal{G}$ into a third-order tensor $\mathcal{G}_r \in \mathbb{R}^{64 \times N \times K}$ by vectorizing its frontal slices, and impose the low rank constraint only to its grouping dimension. We notice that the block matrix representation in (\ref{bcirc_matrix}) may alleviate the unbalance issue, which may further account for the superiority of low rank t-SVD based methods \cite{zhou2017outlier, lu2016tensor}. However, choosing the multi-rank of f-diagonal tensor (\ref{block_diag_tensor}) in the Fourier domain is not easy. In this paper, we adopt the simple hard-thresholding technique to achieve tensor sparsity.

\subsection{Global basis and local group representation}
\begin{figure}[htbp]
		\ssmall
        \tikzstyle{format_my} = [rectangle, rounded corners, minimum width = 0.6cm, minimum height=0.6cm, draw, fill = white]
		\tikzstyle{format}=[rectangle,draw,thin,fill=white]
		\tikzstyle{test}=[diamond,aspect=2,draw,thin]
		\tikzstyle{point}=[coordinate,on grid]
		\begin{tikzpicture}
		\node[format_my] (Input){Noisy Image $\mathcal{A}$};
        \node[format_my, right of=Input, yshift=-0.1cm, below of=Input, xshift=0.2cm, node distance=10.8mm](Ref_patch){$\mathcal{P}_{ref}$};
        \node[format_my, right of=Ref_patch, node distance=16mm] (Grouping){$\mathcal{G}$};
		\node[format_my,right of=Input,yshift=0.1cm, above of=Input, xshift=0.2cm,node distance=10.8mm] (Global_training){$\mathcal{U}_{row}, \mathcal{U}_{col}$};
        \node[format_my,right of=Grouping, node distance=26mm] (Local_representation){$\mathbf{U}_{group}$};
        \node[format_my, right of=Input, yshift=0.3cm, below of=Global_training, xshift=0.1cm, node distance=14.8mm] (Coefficient){$\mathcal{C}$};
        \node[format_my,right of=Coefficient, node distance=28mm] (Filtered_group){$\mathcal{G}_{filtered}$};
        \node[format_my,above of=Filtered_group, node distance=20mm] (Filtered_image){Filtered Image $\mathcal{A}_{c}$};
		\draw[->] (Input)--node[anchor=east]{Global patch basis}(Global_training);
        \draw[->] (Input)--node[anchor=east]{Reference patch}(Ref_patch);
        \draw[->] (Ref_patch)--node[above]{Grouping}(Grouping);
        \draw[->] (Grouping)--node[above]{Local group basis}(Local_representation);
        \draw[->] (Grouping)--(Coefficient);
        \draw[->] (Global_training)--(Coefficient);
        \draw[->] (Local_representation)--(Coefficient);
        \draw[->] (Coefficient)--node[above]{Hard-threshold}node[below]{Inverse transform}(Filtered_group);
        \draw[->] (Filtered_group)--node[anchor=east]{Aggregation}(Filtered_image);
		\end{tikzpicture}
\caption{Implementation of MSt-SVD.}

\label{illus_MSt-SVD}
\end{figure}
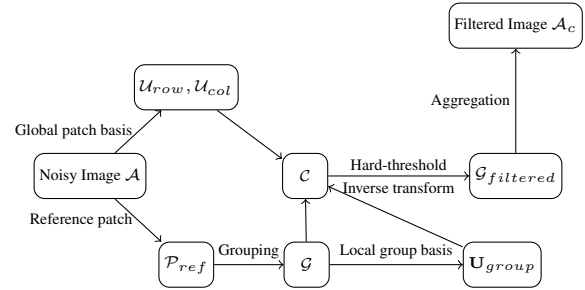

The presence of noise in the training process could distort the local representation and introduce some unwanted artifacts. Some recent denoising strategies \cite{talebi2014global, shikkenawis20162d} consider information of the whole image and learn a global representation to render more robustness. CBM3D shares the same idea by applying a pre-defined transform for all patches. t-SVD is also a suitable alternative as a global basis because it preserves the spatial information, and the pre-defined FFT transform along the third mode could make it less sensitive to the variation of noise. The global patch representation can be trained with randomly sampled patches, but for simplicity, all the reference patches are used.\\
\indent If the patch representation is acquired, then grouping and collaborative filtering could be viewed as a feature extraction and patch classification process that takes care of nonlocal similarity. Therefore, some simple and effective classification method such as PCA can be utilized. We term the combination of global t-SVD basis and local PCA transform as 'multispectral t-SVD (MSt-SVD)', and detailed implementation is given in Algorithm \ref{MSt-SVD} and Fig. \ref{illus_MSt-SVD}. Comparing the FFT matrix $\mathbf{W}$ in (\ref{fft_matrix}) and the opponent color mode transform matrix $\mathbf{U}_{color}^T$ of CBM3D in (\ref{opp_color_mode}), it can be seen that the first slice in the Fourier domain corresponding to the first row of $\mathbf{W}$ can be regarded as luminance channel, thus similar to CBM3D, the grouping process and the training of local PCA can be performed by considering only the first slices of all patches in the Fourier domain. Obviously, this implementation can save $\frac{2}{3}$ computational time on grouping and training. This efficient modification of MSt-SVD for color image is termed 'color MSt-SVD' (CMSt-SVD), and its implementation is briefed in Algorithm \ref{CMSt-SVD}.\\
\indent Interestingly, the relationship among nonlocal t-SVD, CBM3D and 4DHOSVD is therefore established using block diagonal representation in the Fourier domain.

\begin{algorithm}[ht]
\caption{MSt-SVD} 
{\bf Input:} Color or Multispectral image $\mathcal{A}$, patch size $ps$, local search window size $SR$, number of similar patches $K$, pixels between two adjacent reference patches $N_{step}$.\\
{\bf Output:} Filtered image $\mathcal{A}_c$.\\
{\bf Step 1} (Global Training): Train the global patch representation $\mathcal{U}_{row}$ and $\mathcal{U}_{column}$ with all reference patches using the nonlocal t-SVD in (\ref{nonlocal_tSVD}).\\
{\bf Step 2} (Grouping): Given reference patch $\mathcal{P}_{ref}$, calculate its Euclidean distance with all patches located in $SR$ via $||\mathcal{P}_{ref} - \mathcal{P}_{i}||_F$ to stack $K$ most similar patches in a group $\mathcal{G}$.\\
{\bf Step 3} (Collaborative filtering):\\
 \hspace*{0.2in}(1) Learn a factor matrix $\mathbf{U}_{group}$ in the 4-th mode of $\mathcal{G}$ via full PCA, and obtain the core tensor $\mathcal{C}$ via (\ref{compute_C_4D}).\\
 \hspace*{0.2in}(2) Apply the hard-threshold technique to $\mathcal{C}$, whose elements smaller than a certain threshold is set to zero.\\
 \hspace*{0.2in}(3) Obtain filtered group $\mathcal{G}_{filtered}$ via (\ref{compute_filtered_group}).\\
{\bf Step 4} (Aggregation): Averagely write back all image patches in $\mathcal{G}_d$ to their original locations.
\label{MSt-SVD}
\end{algorithm}

\begin{algorithm}
\caption{CMSt-SVD} 
{\bf Input:} Noisy color image $\mathcal{A}$, $ps$, $SR$, $K$ and $N_{step}$.\\
{\bf Output:} Filtered image $\mathcal{A}_c$.\\
{\bf Step 1} (Global Training): The same as MSt-SVD.\\
{\bf Step 2} (Grouping): Given reference patch $\mathcal{P}_{ref}$, calculate its Euclidean distance with all patches in the Fourier domain using only the first slices via $||\hat{\mathcal{P}}_{ref}(:,:,1) - \hat{\mathcal{P}}_{i}(:,:,1)||_F$ to stack $K$ most similar patches in a group $\mathcal{G}$.\\
{\bf Step 3} (Collaborative filtering):\\
 \hspace*{0.2in}(1) Learn a factor matrix $\mathbf{U}_{group}$ in the last mode of third order tensor $\hat{\mathcal{G}}(:,:,1,:)$ in the Fourier domain via full PCA, and obtain the core tensor $\mathcal{C}$ via (\ref{compute_C_4D}).\\
 \hspace*{0.2in}(2) and (3) are the same as MSt-SVD.\\
{\bf Step 4} (Aggregation): The same as MSt-SVD.
\label{CMSt-SVD}
\end{algorithm}

\subsection{Computational complexity}
In this subsection, we compare the computational complexity of CBM3D, 4DHOSVD and the proposed MSt-SVD. For simplicity, we assume that the number of image pixels is $N$, that the average time to compute similar patches per reference patch is $T_{s}$, that the average number of patches similar to the reference patch is $K$, and that the size of the patch is $p\times p$ ($p\ll K$). According to \cite{rajwade2013image}, the time complexity of 4DHOSVD and CBM3D are $O([T_s+Kp^3+Kp^4]N)$ and $O([T_s+Kp^2logp+p^2KlogK]N)$, respectively. The computational burden of MSt-SVD mainly lies in the PCA transform $O(Kp^4)$ and patch level t-SVD transform $O(Kp^3)$, leading to a total complexity of $O([T_s+Kp^3+Kp^4]N)$. Considering that MSt-SVD is a one step algorithm and does not require the training of patch level transform for each group, it is competitive in terms of efficiency.

\section{Experiments}
In this section, we evaluate the performance of MSt-SVD and CMSt-SVD for color image and multispectral image denoising. All the results of compared methods are obtained by fine-tuned parameters or from the authors' papers. All the experiments are performed on a moderate laptop equipped with Core(TM) i5-8250U @ 1.8 GHz and 8GB RAM. Our software package is publicly available\footnote{https://github.com/ZhaomingKong}, which includes a fast C++ mex-function that could help reproduce all our results of color image and multispectral image denoising within 10 and 100 seconds, respectively.
\subsection{Experimental setting for color image}
\indent A brief description of four publicly available real-world datasets is listed in Table \ref{Table_dataset_description}, and more detailed information is in \cite{DatasetXu} and \cite{KongData}.
\begin{table}[htbp]
  \centering
  \caption{Brief description of three real-world color image datasets.}
    \begin{tabular}{cccc}
    \toprule
    Dataset & Type  & \# of Images & Image Size \\
    \midrule
    CC15 \cite{nam2016holistic}  & Real-World & 15    & $512 \times 512$ \\
    \midrule
    CC60  \cite{xu2018external} & Real-World & 60    & $500 \times 500$ \\
    \midrule
    Xu-100    \cite{DatasetXu} & Real-World & 100   & $512 \times 512$ \\
    \midrule
    Ours    \cite{KongData} & Real-World & 249   & $1024 \times 1024$ \\
    \bottomrule
    \end{tabular}%
  \label{Table_dataset_description}%
\end{table}%

\indent The representative compared methods for color image denoising include: CBM3D \cite{dabov2007color}, 4DHOSVD1 (hard-thresholding) \cite{rajwade2013image}, WTR1 \cite{Li2018Weighted}, Neat Image (NI), TNRD \cite{Chen2015On}, GID \cite{xu2018external}, MCWNNM \cite{xu2017multi}, TWSC \cite{TrilateralXu}, LSCD \cite{rizkinia2016local}, and LLRT \cite{chang2017hyper}. Three representative neural network based methods MLP \cite{burger2012image}, DnCNN \cite{zhang2017beyond} and FFD-Net \cite{Zhang2018FFDNet} are also included in our comparison. Considering the computational complexity of some compared methods and for fair comparison, all methods are tuned to produce their best average results, and the input noise level $\sigma$ of some Gaussian denoisers is listed in Table \ref{Table_sigma_chosen}. In practical implementation, however, $\sigma$ should be tuned for every image, so to better understand the effectiveness of state-of-the-art CBM3D, the best result of CBM3D on every image is reported, and this implementation is termed 'CBM3D\_{best}'. PSNR and SSIM indices are employed for objective evaluation.
\begin{table}[htbp]
  \centering
  \caption{Input noise level of Gaussian denoisers for CC15, CC60 and Xu-100 datasets.}
    \begin{tabular}{ccccc}
    \toprule
    Method & LSCD  & WTR1  & FFD-Net & LLRT \\
    \midrule
    $\sigma$ & 10    & 15    & 15    & 20 \\
    \midrule
    Method & 4DHOSVD1 & CBM3D & MSt-SVD & CMSt-SVD \\
    \midrule
    $\sigma$ & 30    & 20    & 30    & 25 \\
    \bottomrule
    \end{tabular}%
  \label{Table_sigma_chosen}%
\end{table}%

\subsection{Experimental results for color image}
\subsubsection{Experiments on CC15}
\begin{figure*}[htbp]
\graphicspath{{Illustration/Compare_dataset_CC15/Combine_new/}}
\centering
\subfigure[Clean]{
\label{Fig4}
\includegraphics[width=1.08in]{Clean_demo1_combine_paint}}
\subfigure[Noisy]{
\label{Fig4}
\includegraphics[width=1.08in]{Noisy_demo1_combine}}
\subfigure[FFD-Net]{
\label{Fig4}
\includegraphics[width=1.08in]{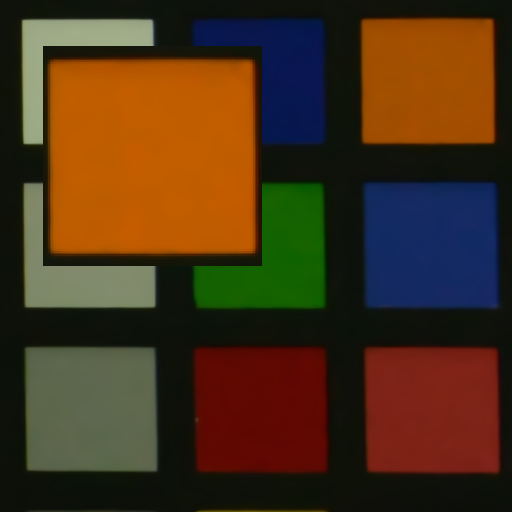}}
\subfigure[LLRT]{
\label{Fig5}
\includegraphics[width=1.08in]{LLRT_demo1_combine}}
\subfigure[GID]{
\label{Fig5}
\includegraphics[width=1.08in]{GID_demo1_combine}}\\

\subfigure[TWSC]{
\label{Fig4}
\includegraphics[width=1.08in]{TWSC_demo1_combine}}
\subfigure[MCWNNM]{
\label{Fig4}
\includegraphics[width=1.08in]{MCWNNM_demo1_combine}}
\subfigure[4DHOSVD1]{
\label{Fig4}
\includegraphics[width=1.08in]{4DHOSVD1_demo1_combine}}
\subfigure[CBM3D]{
\label{Fig4}
\includegraphics[width=1.08in]{CBM3D_demo1_combine}}
\subfigure[CMSt-SVD]{
\label{Fig4}
\includegraphics[width=1.08in]{MS-TSVD_demo1_combine}}

\caption{Denoised images of compared methods on CC15. The camera is CANON D600 with ISO = 3200. Please zoom-in for better view.}

\label{Fig_illus_CC15_demo1}
\end{figure*}

\begin{figure*}[htbp]
\graphicspath{{Illustration/Compare_dataset_CC15/Combine_new/}}
\centering
\subfigure[Clean]{
\label{Fig4}
\includegraphics[width=1.08in]{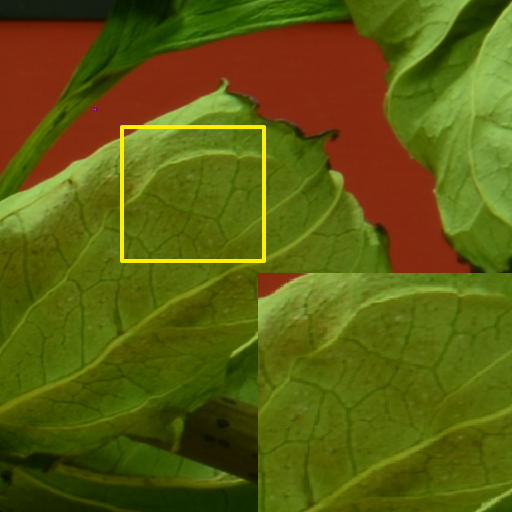}}
\subfigure[Noisy]{
\label{Fig4}
\includegraphics[width=1.08in]{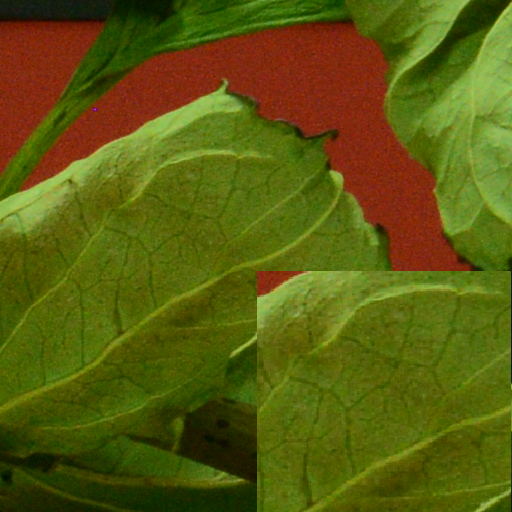}}
\subfigure[FFD-Net]{
\label{Fig4}
\includegraphics[width=1.08in]{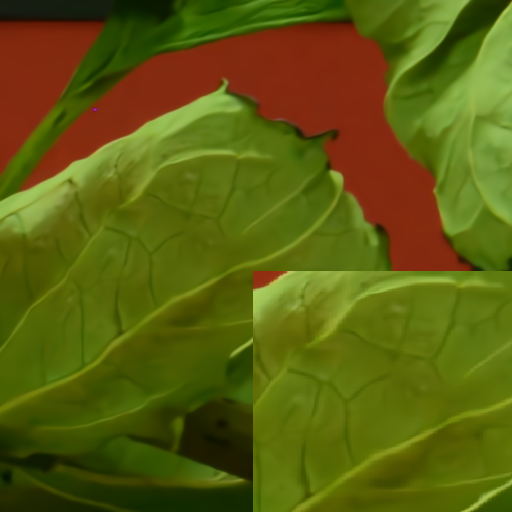}}
\subfigure[LLRT]{
\label{Fig5}
\includegraphics[width=1.08in]{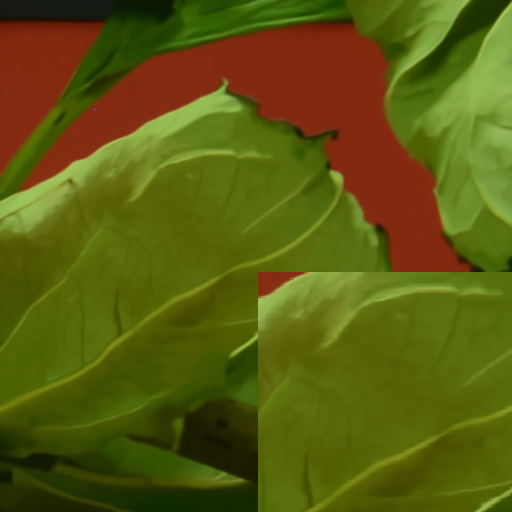}}
\subfigure[GID]{
\label{Fig5}
\includegraphics[width=1.08in]{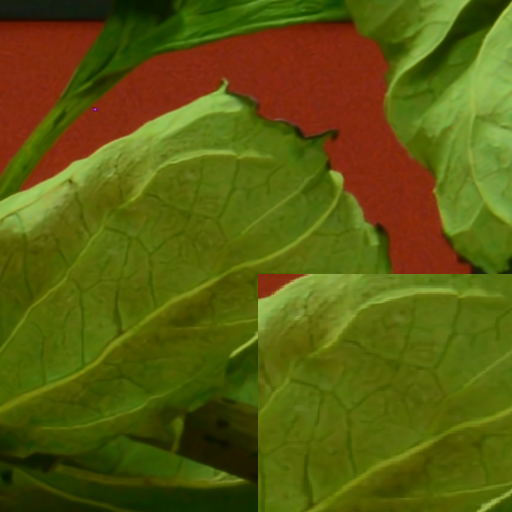}}\\

\subfigure[TWSC]{
\label{Fig4}
\includegraphics[width=1.08in]{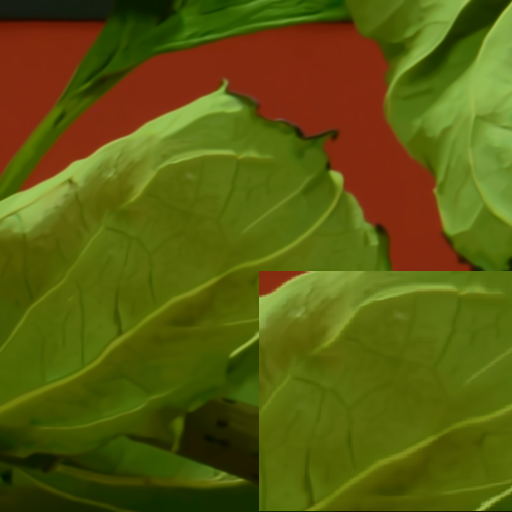}}
\subfigure[MCWNNM]{
\label{Fig4}
\includegraphics[width=1.08in]{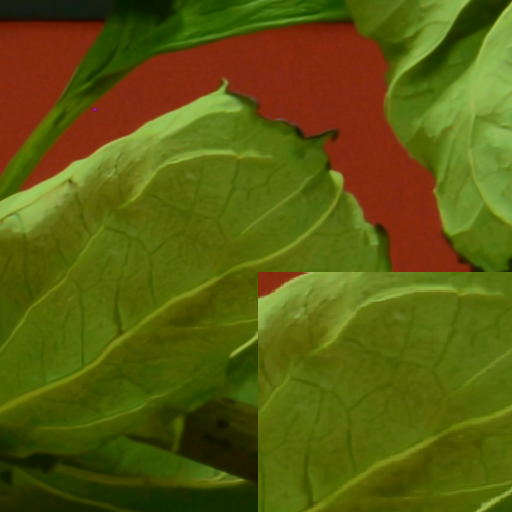}}
\subfigure[4DHOSVD1]{
\label{Fig4}
\includegraphics[width=1.08in]{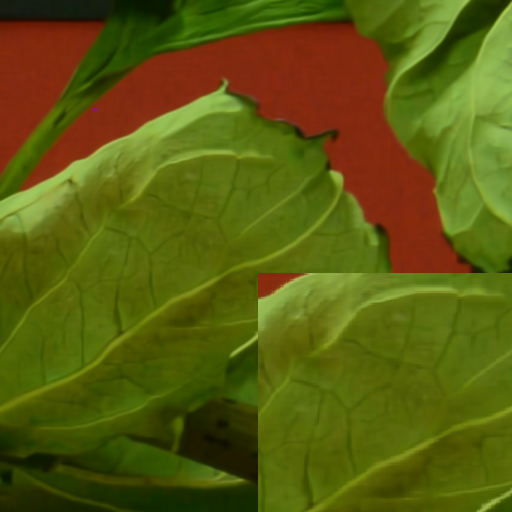}}
\subfigure[CBM3D]{
\label{Fig4}
\includegraphics[width=1.08in]{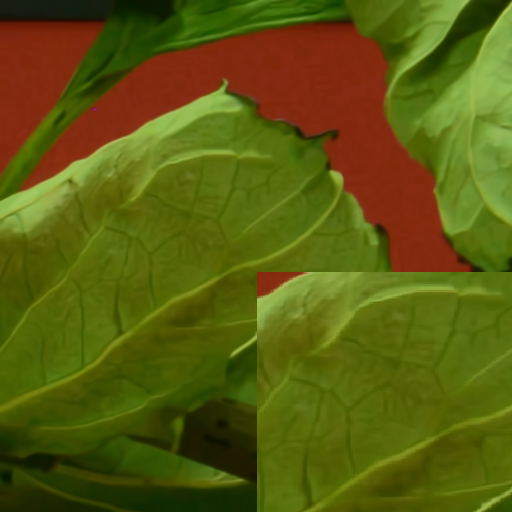}}
\subfigure[CMSt-SVD]{
\label{Fig4}
\includegraphics[width=1.08in]{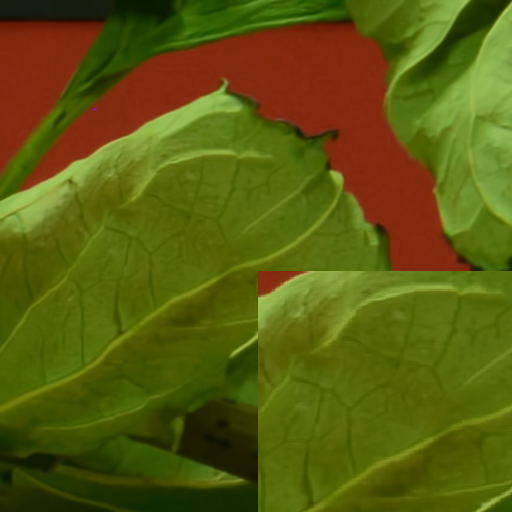}}

\caption{Denoised images of compared methods on CC15. The camera is NIKON D800 with ISO = 1600. Please zoom-in for better view.}

\label{Fig_illus_CC15_demo2}
\end{figure*}

The fine-tuned results of TNRD, MLP and DnCNN in \cite{xu2017multi} are used. PSNR result of every image and average computational time are listed in Table \ref{Table_cc15}\footnote{CBM3D and LSCD use a C++ mex function, while other methods are implemented purely with Matlab.}, and visual evaluations are given in Fig. \ref{Fig_illus_CC15_demo1} and Fig. \ref{Fig_illus_CC15_demo2}. Table \ref{Table_cc15} shows that the simple CMSt-SVD consistently outperforms MSt-SVD and 4DHOSVD1, and both MSt-SVD and CMSt-SVD are able to produce very competitive performance in terms of both effectiveness and efficiency. Fig. \ref{Fig_illus_CC15_demo1} shows that the representative low-rank based method LLRT and the sparse coding scheme TWSC produce satisfactory results in homogenous regions, because the underlying clean patches share much similar feature, and thus can be modeled as a low-rank or sparse coding problem. But as illustrated in Fig. \ref{Fig_illus_CC15_demo2}, when the ground truth image contains more details or local variations, clear over-smooth effects can be observed. Interestingly, the state-of-the-art neural network FFD-Net shows similar effects. Besides, compared with CBM3D and CMSt-SVD, the local 4DHOSVD transform is more easily affected by the presence of noise, which is incorporated in the training process of color mode transform.

\begin{table*}[htbp]
  \centering
  \ssmall
  \caption{PSNR results and average computational time (s) of different methods on CC15 dataset. The three best results are bolded.}
    \begin{tabular}{cccccccccccccccccc}
    \toprule
    Methods & \multicolumn{3}{c}{Canon 5D ISO = 3200} & \multicolumn{3}{c}{Canon D600 ISO = 3200} & \multicolumn{3}{c}{Nikon D800 ISO = 1600} & \multicolumn{3}{c}{Nikon D800 ISO = 3200} & \multicolumn{3}{c}{Nikon D800 ISO = 6400} & Average & Time (s) \\
    \midrule
    LSCD  & 37.86  & 36.21  & 35.52  & 34.65  & 36.26  & 38.24  & 37.90  & 38.88  & 38.32  & 37.45  & 36.49  & 37.73  & 32.33  & 32.55  & 32.62  & 36.20  & 9.68  \\
    \midrule
    LLRT  & 39.23  & 36.31  & 35.93  & 34.74  & 36.83  & 40.58  & 37.39  & 40.27  & 37.78  & 39.79  & 37.34  & 41.03  & 35.09  & 34.05  & 34.11  & 37.36  & $>1000$ \\
    \midrule
    WTR1  & 41.09  & 36.92  & 36.25  & 34.68  & 36.48  & 40.52  & 38.26  & 41.40  & 38.61  & 39.98  & 37.70  & 41.36  & 35.16  & 34.22  & 34.43  & 37.81  & $>2000$ \\
    \midrule
    TNRD  & 39.51  & 36.47  & 36.45  & 34.79  & 36.37  & 39.49  & 38.11  & 40.52  & 38.17  & 37.69  & 35.90  & 38.21  & 32.81  & 32.33  & 32.29  & 36.61  & N/A \\
    \midrule
    MLP   & 39.00  & 36.34  & 36.33  & 34.70  & 36.20  & 39.33  & 37.95  & 40.23  & 37.94  & 37.55  & 35.91  & 38.15  & 32.69  & 32.33  & 32.29  & 36.46  & N/A \\
    \midrule
    DnCNN & 37.26  & 34.13  & 34.09  & 33.62  & 34.48  & 35.41  & 35.79  & 36.08  & 35.48  & 34.08  & 33.70  & 33.31  & 29.83  & 30.55  & 30.09  & 33.86  & N/A \\
    \midrule
    GID   & 40.82  & 37.19  & 36.92  & 35.32  & 36.62  & 38.68  & 38.88  & 40.66  & 39.20  & 37.92  & 36.62  & 37.64  & 33.01  & 32.93  & 32.96  & 37.02  & 55.60  \\
    \midrule
    WCMNNM & 41.20  & 37.25  & 36.48  & 35.54  & 37.03  & 39.56  & 39.26  & 41.45  & 39.54  & 38.94  & 37.40  & 39.42  & 34.85  & 33.97  & 33.97  & 37.72  & 318.29  \\
    \midrule
    FFD-Net & 39.40  & 37.02  & 36.53  & 34.97  & 36.73  & 41.02  & 38.66  & 41.53  & 38.80  & 40.15  & 37.61  & 41.18  & 34.13  & 33.66  & 33.69  & 37.68  & 28.98  \\
    \midrule
    TWSC  & 40.55  & 35.92  & 35.15  & 35.36  & 37.09  & 41.13  & 39.36  & 41.91  & 38.81  & 40.27  & 37.22  & 42.09  & 35.53  & 34.15  & 33.93  & \textbf{37.90 } & 480.80  \\
    \midrule
    CBM3D & 40.77  & 37.31  & 36.98  & 35.21  & 36.76  & 40.13  & 39.02  & 41.65  & 39.40  & 39.59  & 37.49  & 39.47  & 34.13  & 33.73  & 33.85  & 37.69  & 6.98  \\
    \midrule
    CBM3D\_best & 40.96  & 37.31  & 37.15  & 35.38  & 36.81  & 40.45  & 39.25  & 41.65  & 39.59  & 39.86  & 37.54  & 40.38  & 34.85  & 33.92  & 34.16  & \textbf{37.95 } & 6.98  \\
    \midrule
    4DHOSVD1 & 40.22  & 36.97  & 36.55  & 35.02  & 36.60  & 39.78  & 38.85  & 41.35  & 39.11  & 39.24  & 37.28  & 39.47  & 34.40  & 33.81  & 34.01  & 37.51  & 120.18  \\
    \midrule
    MSt-SVD & 40.33  & 37.25  & 36.83  & 35.16  & 36.71  & 40.29  & 38.97  & 41.49  & 39.24  & 39.61  & 37.43  & 39.93  & 34.34  & 33.82  & 33.96  & 37.69  & 110.06  \\
    \midrule
    CMSt-SVD & 40.79  & 37.37  & 37.01  & 35.29  & 36.95  & 40.93  & 39.21  & 41.98  & 39.54  & 39.98  & 37.65  & 40.05  & 34.50  & 33.93  & 34.01  & \textbf{37.95 } & 98.88  \\
    \bottomrule
    \end{tabular}%
  \label{Table_cc15}
\end{table*}%

\subsubsection{Experiments on CC60, Xu-100 and our datasets}
\begin{figure*}[htbp]
\graphicspath{{Illustration/Compare_dataset_CC60/Combine_new/}}
\centering
\subfigure[Clean]{
\label{Fig4}
\includegraphics[width=1.11in]{Clean_demo1_combine_paint}}
\subfigure[Noisy]{
\label{Fig4}
\includegraphics[width=1.11in]{Noisy_demo1_combine}}
\subfigure[NI]{
\label{Fig4}
\includegraphics[width=1.11in]{NI_demo1_combine}}
\subfigure[LLRT]{
\label{Fig5}
\includegraphics[width=1.11in]{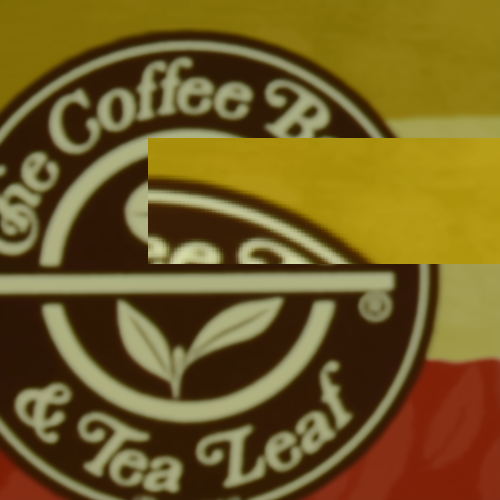}}
\subfigure[GID]{
\label{Fig5}
\includegraphics[width=1.11in]{GID_demo1_combine}}\\

\subfigure[TWSC]{
\label{Fig4}
\includegraphics[width=1.11in]{TWSC_demo1_combine}}
\subfigure[MCWNNM]{
\label{Fig4}
\includegraphics[width=1.11in]{MCWNNM_demo1_combine}}
\subfigure[4DHOSVD1]{
\label{Fig4}
\includegraphics[width=1.11in]{4DHOSVD1_demo1_combine}}
\subfigure[CBM3D]{
\label{Fig4}
\includegraphics[width=1.11in]{CBM3D_demo1_combine}}
\subfigure[CMSt-SVD]{
\label{Fig4}
\includegraphics[width=1.11in]{MS-TSVD_demo1_combine}}

\caption{Denoised images of compared methods on CC60. The camera is NIKON D800 with ISO = 1600. Please zoom-in for better view.}

\label{Fig_illus_CC60}
\end{figure*}

\begin{figure*}[htbp]
\graphicspath{{Illustration/Compare_dataset_Xu/Combine_new/}}
\centering
\subfigure[Clean]{
\label{Fig4}
\includegraphics[width=1.11in]{Clean_demo1_combine_paint}}
\subfigure[Noisy]{
\label{Fig4}
\includegraphics[width=1.11in]{Noisy_demo1_combine}}
\subfigure[NI]{
\label{Fig4}
\includegraphics[width=1.11in]{NI_demo1_combine}}
\subfigure[LLRT]{
\label{Fig5}
\includegraphics[width=1.11in]{LLRT_demo1_combine}}
\subfigure[GID]{
\label{Fig5}
\includegraphics[width=1.11in]{GID_demo1_combine}}\\

\subfigure[TWSC]{
\label{Fig4}
\includegraphics[width=1.11in]{TWSC_demo1_combine}}
\subfigure[MCWNNM]{
\label{Fig4}
\includegraphics[width=1.11in]{MCWNNM_demo1_combine}}
\subfigure[4DHOSVD1]{
\label{Fig4}
\includegraphics[width=1.11in]{4DHOSVD1_demo1_combine}}
\subfigure[CBM3D]{
\label{Fig4}
\includegraphics[width=1.11in]{CBM3D_demo1_combine}}
\subfigure[CMSt-SVD]{
\label{Fig4}
\includegraphics[width=1.11in]{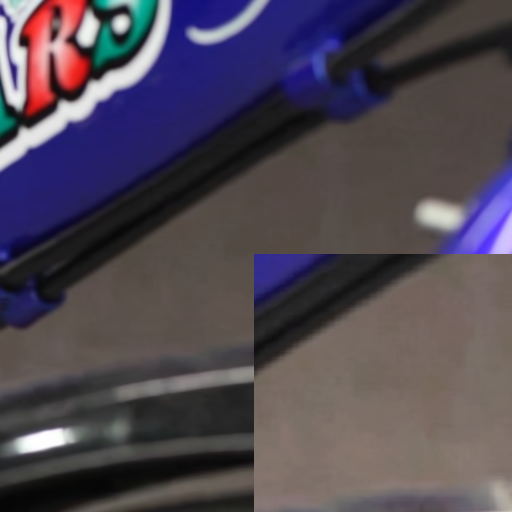}}

\caption{Dennoised images of compared methods on Xu's dataset. The camera is CANON 5D with ISO = 6400. Please zoom-in for better view.}

\label{Fig_illus_Xu}
\end{figure*}

\begin{figure}[htbp]
\graphicspath{{Illustration/Compare_my_own/new/}}
\centering
\subfigure[Clean]{
\label{Fig4}
\includegraphics[width=0.79in]{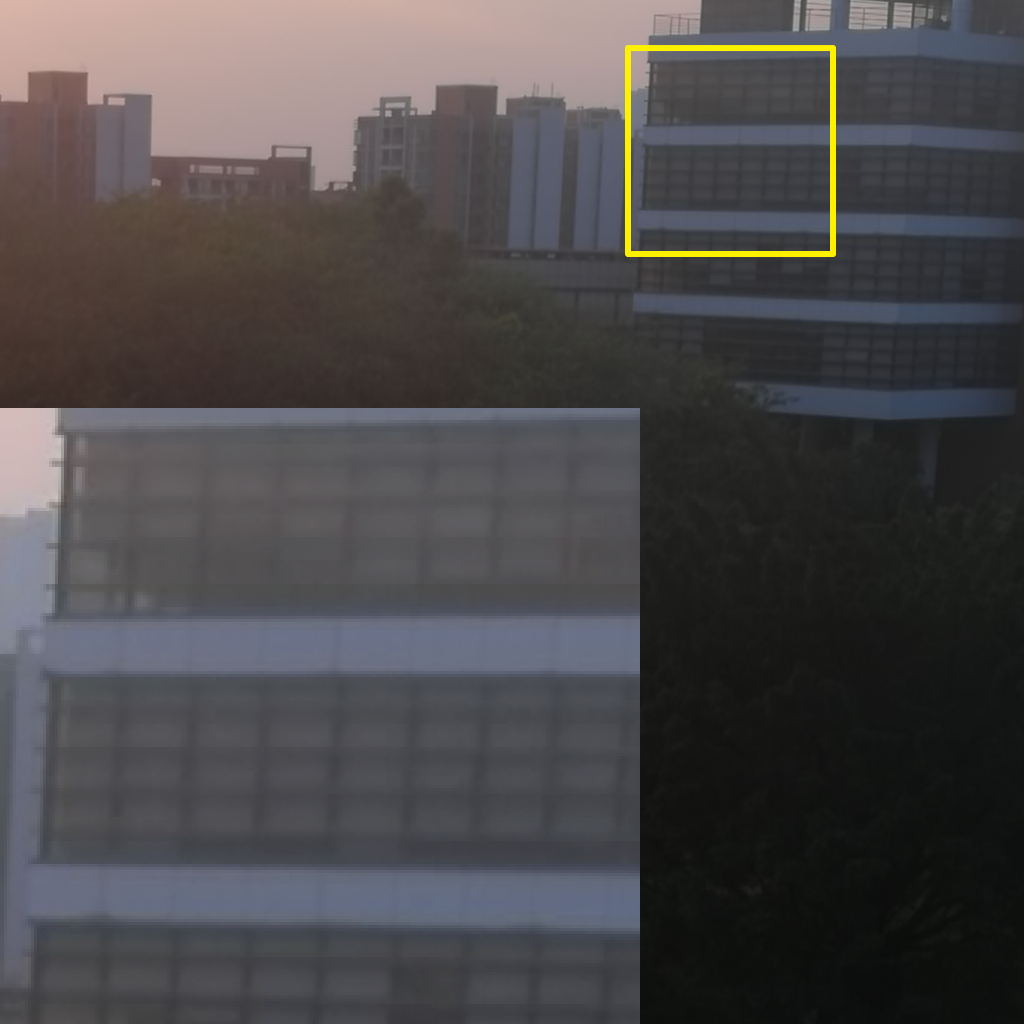}}
\subfigure[Noisy]{
\label{Fig4}
\includegraphics[width=0.79in]{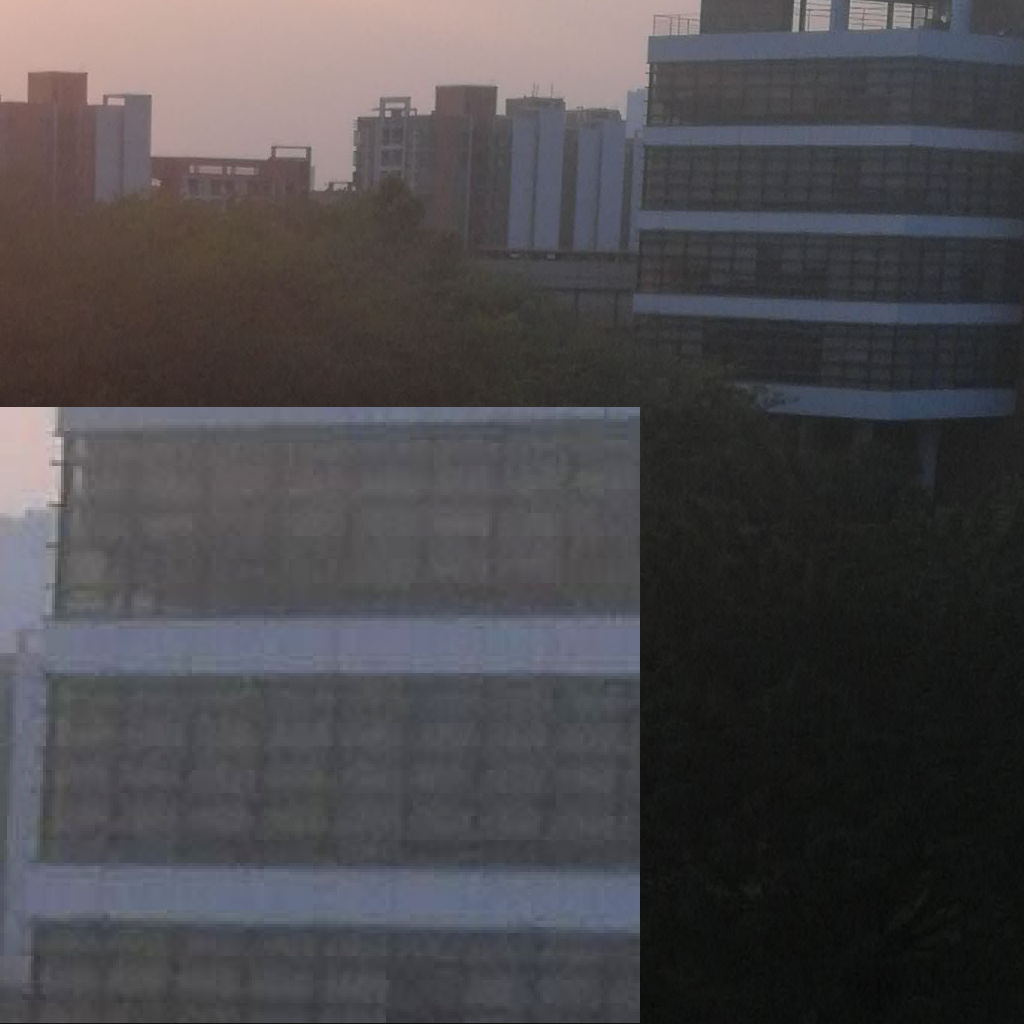}}
\subfigure[FFD-Net]{
\label{Fig4}
\includegraphics[width=0.79in]{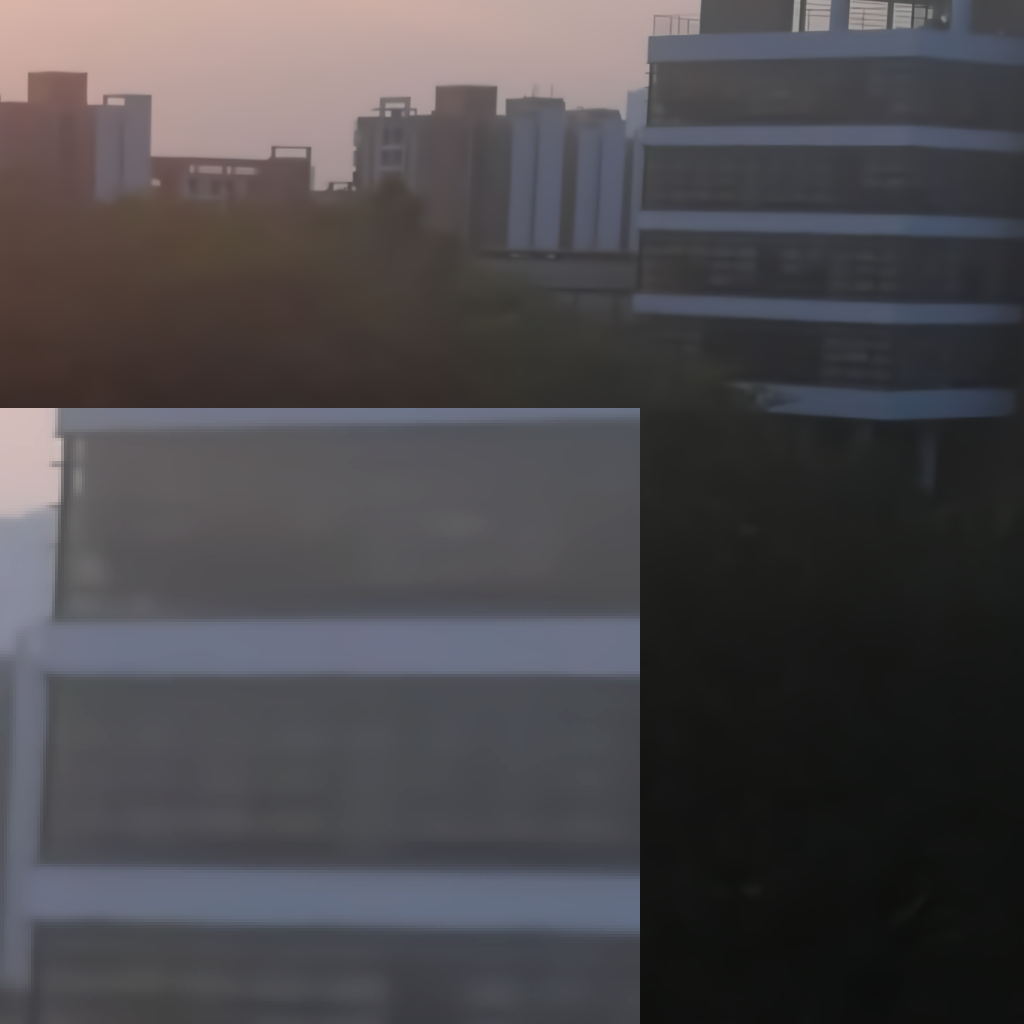}}
\subfigure[CMSt-SVD]{
\label{Fig5}
\includegraphics[width=0.79in]{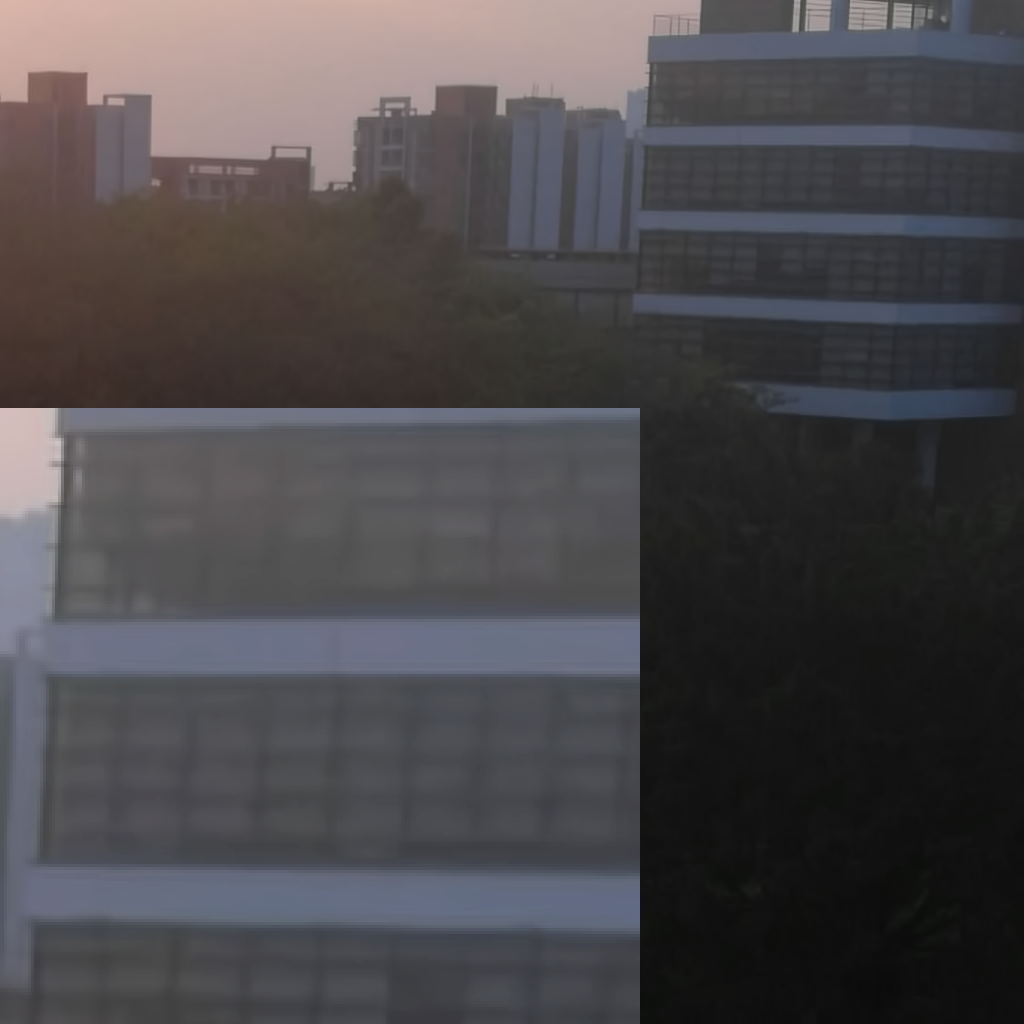}}

\caption{Denoised images of FFD-Net and CMSt-SVD on our dataset. The camera is HUAWEI Honor 6X with auto mode. Please zoom-in for better view.}

\label{Fig_my_own}
\end{figure}

Table \ref{Table_CC60_Xu} and \ref{Table_my_own} list the average PSNR and SSIM values of several competitive methods. It is obvious that CMSt-SVD still demonstrates one of the best performance. Our visual evaluations in Fig. \ref{Fig_illus_CC60} and Fig. \ref{Fig_my_own} further illustrate the over-smooth effects of the sparse coding scheme and state-of-the-art neural network, while Fig. \ref{Fig_illus_Xu} shows that many competitive methods including NI, 4DHOSVD, MCWNNM and GID produce color artifacts to some degree, which is similar to our observation in Fig. \ref{Fig_illus_CC15_demo1}.
\begin{table*}[htbp]
\scriptsize
  \centering
  \caption{Average PSNR and SSIM values of compared methods on CC60 and Xu-100 datasets. The best results are bolded.}
    \begin{tabular}{cccccccccccc}
    \toprule
    Dataset & Index & LLRT  & WTR1  & FFD-Net & GID   & TWSC  & MCWNNM & 4DHOSVD1 & CBM3D & CBM3D\_best & CMSt-SVD \\
    \midrule
    \multirow{2}[4]{*}{CC60} & PSNR  & 38.51  & 39.69  & \textbf{39.73 } & 38.41  & 39.66  & 39.03  & 39.15  & 39.40  & 39.68  & \textbf{39.75 } \\
\cmidrule{2-12}          & SSIM  & 0.9636  & 0.9764  & 0.9770  & 0.9633  & 0.9759  & 0.9698  & 0.9729  & 0.9740  & \textbf{0.9775 } & 0.9756  \\
    \midrule
    \multirow{2}[4]{*}{Xu} & PSNR  & 38.51 & 38.56 & 38.56  & 38.37 & 38.62 & 38.51 & 38.51 & 38.69 & \textbf{38.81} & \textbf{38.82} \\
\cmidrule{2-12}          & SSIM  & 0.9707 & 0.9669 & 0.9658  & 0.9675 & 0.9674 & 0.9671 & 0.9673 & 0.9694 & \textbf{0.9712} & 0.9694 \\
    \bottomrule
    \end{tabular}%
  \label{Table_CC60_Xu}%
\end{table*}%

\begin{table*}[htbp]
\ssmall
  \centering
  \caption{Average PSNR and SSIM values of compared methods on our dataset. The best results are bolded. The input noise level of compared methods are tuned to produce their best average results of every camera.}
    \begin{tabular}{cccccccccccc}
    \toprule
    Camera & \# of Images & Index & LLRT  & FFD-Net & GID   & TWSC  & MCWNNM & 4DHOSVD1 & CBM3D & CBM3D\_best & CMSt-SVD \\
    \midrule
    \multirow{2}[4]{*}{HUAWEI HONOR 6X} & \multirow{2}[4]{*}{30} & PSNR  & 39.54 & 40.05  & 39.52  & 39.71  & 39.46 & 39.82 & 39.97 & \textbf{40.48} & 40.08 \\
\cmidrule{3-12}          &       & SSIM  & 0.9669 & 0.9669  & 0.9653 & 0.9651 & 0.9610  & 0.9658  & 0.9669  & \textbf{0.9740 } & 0.9674  \\
    \midrule
    \multirow{2}[4]{*}{IPHONE 5S} & \multirow{2}[4]{*}{36} & PSNR  & 40.02 & 40.60  & 40.12  & 40.27  & 39.87 & 40.68 & 40.77 & \textbf{41.25} & 40.84 \\
\cmidrule{3-12}          &       & SSIM  & 0.9676 & 0.9645  & 0.9642 & 0.9617 & 0.9567  & 0.9664  & 0.9668  & \textbf{0.9758 } & 0.9668  \\
    \midrule
    \multirow{2}[4]{*}{IPHONE 6S} & \multirow{2}[4]{*}{67} & PSNR  & 39.72 & 40.49  & 40.16  & 40.12  & 40.18 & 40.36 & 40.55 & \textbf{41.16} & 40.53 \\
\cmidrule{3-12}          &       & SSIM  & 0.9663 & 0.9707  & 0.9670  & 0.9619  & 0.9628  & 0.9671  & 0.9693  & \textbf{0.9783 } & 0.9674  \\
    \midrule
    \multirow{2}[4]{*}{CANON 100D} & \multirow{2}[4]{*}{55} & PSNR  & 41.84 & 41.67  & 40.86 & 41.65 & 41.47 & 41.41 & 41.69 & \textbf{42.08} & 41.99 \\
\cmidrule{3-12}          &       & SSIM  & 0.9784 & 0.9768  & 0.9743  & 0.9767  & 0.9774  & 0.9771  & 0.9780  & \textbf{0.9808 } & 0.9794  \\
    \midrule
    \multirow{2}[4]{*}{CANON 600D} & \multirow{2}[4]{*}{25} & PSNR  & 42.53 & 42.55  & 41.60  & 42.52  & 42.07 & 42.14 & 42.54 & \textbf{42.89} & 42.75 \\
\cmidrule{3-12}          &       & SSIM  & 0.9816 & 0.9824  & 0.9790  & 0.9824  & 0.9795  & 0.9810  & 0.9836  & \textbf{0.9851 } & 0.9840  \\
    \midrule
    \multirow{2}[4]{*}{SONY A6500} & \multirow{2}[4]{*}{36} & PSNR  & 45.71 & 45.71  & 44.94  & 45.48  & 45.37 & 45.56 & 45.70  & 45.81 & \textbf{45.89} \\
\cmidrule{3-12}          &       & SSIM  & 0.9899 & 0.9901  & 0.9887  & 0.9896  & 0.9894  & 0.9901  & 0.9902  & \textbf{0.9904 } & 0.9903  \\
    \bottomrule
    \end{tabular}%
  \label{Table_my_own}%
\end{table*}%

\subsubsection{Visual evaluation on dnd dataset \cite{plotz2017benchmarking}}
\begin{figure}[htbp]
\graphicspath{{Illustration/DND/}}
\centering
\subfigure[Noisy]{
\label{Fig4}
\includegraphics[width=0.79in]{noisy_crop}}
\subfigure[Neat Image]{
\label{Fig4}
\includegraphics[width=0.79in]{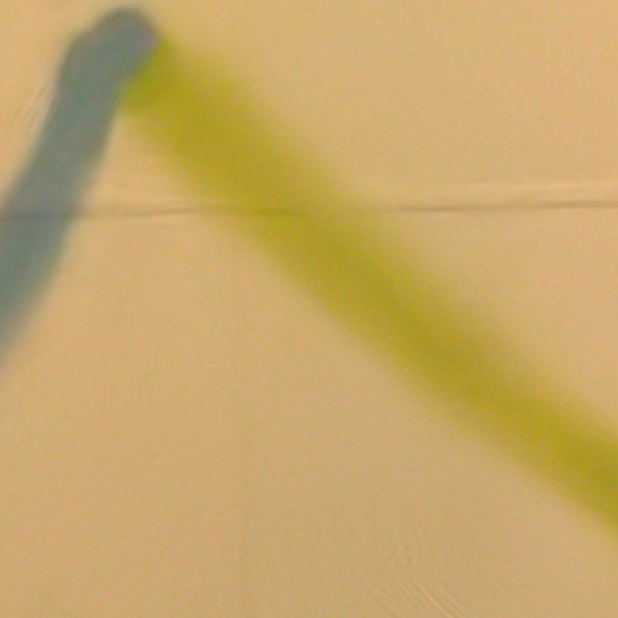}}
\subfigure[CBM3D]{
\label{Fig4}
\includegraphics[width=0.79in]{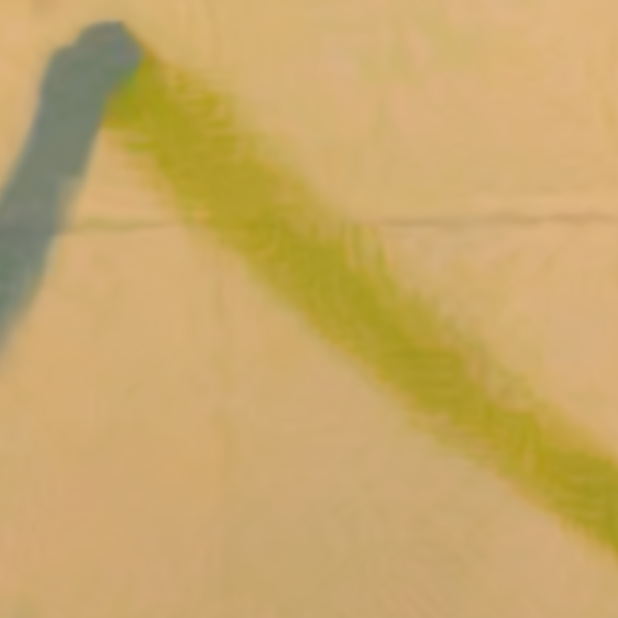}}
\subfigure[MSt-SVD]{
\label{Fig4}
\includegraphics[width=0.79in]{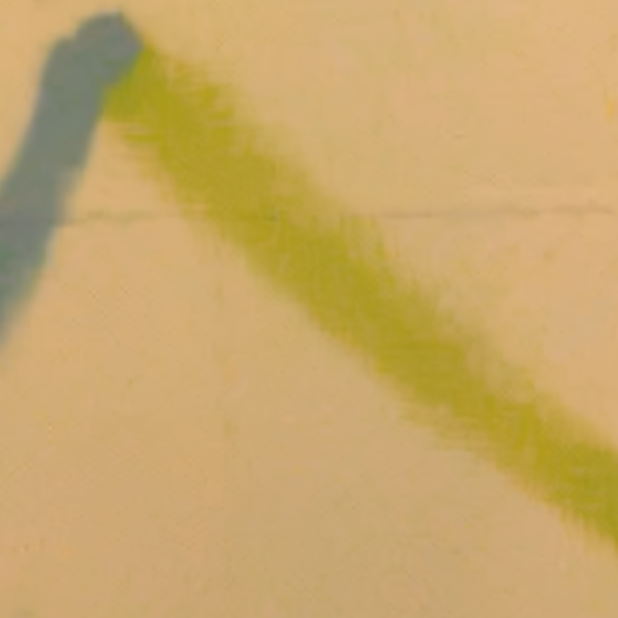}}

\caption{Visual comparison on real dnd dataset.}

\label{fig_real_dnd}
\end{figure}

The ground truth images of this dataset is not available, so we choose one severely corrupted image that contains lines, smooth regions, color texture and details. The input noise level $\sigma$ of MSt-SVD\footnote{An efficient tool is available at github.com/ZhaomingKong/Pure\_Image} and CBM3D is tuned to achieve the best possible visual effects, and about $\sigma = 50 $ is used for both MSt-SVD and CBM3D. Also, the parameters of commercial software Neat Image are carefully chosen to compare their difference. Fig. \ref{fig_real_dnd} shows the visual comparison. Unfortunately, all compared methods introduce unwanted artifacts. Neat Image presents the best results at line areas, while MSt-SVD produces sharper details with the green color uniformly distributed. The benchmark CBM3D with predefined transforms seeks a balance between details and smoothness.

\subsection{Experimental setting for multispectral image}
The representative compared methods for multispectral image denoising include: LRTA \cite{renard2008denoising}, PARAFAC \cite{liu2012denoising}, 4DHOSVD1, LLRT, BM4D \cite{Maggioni2013Nonlocal}, TDL \cite{peng2014decomposable}, ISTReg \cite{xie2016multispectral}, LRMR \cite{zhang2014hyperspectral} and Nmog \cite{chen2018denoising}. \\
\indent Four quality indices are employed for multispectral image: PSNR, SSIM, ERGAS \cite{wald2002data} and SAM \cite{yuhas1993determination}. EGRAS and SAM are spectral-based evaluation indices, and the smaller EGRAS and SAM values are, the better the restored images are. \\
\indent Three publicly available datasets are used: Columbia Multispectral Dataset (CAVE)\footnote{www1.cs.columbia.edu/CAVE/databases/multispectral} for simulated experiments, Harvard Hyperspectral Dataset (HHD)\footnote{vision.seas.harvard.edu/hyperspec/index.html} \cite{chakrabarti2011statistics} and Urban\footnote{www.tec.army.mil/hypercube} dataset for the real cases.
\subsection{Experimental results for Multispectral image}
\subsubsection{i.i.d. Gaussian Noise}
\begin{table*}[htbp]
\ssmall
  \centering
  \caption{Quantitative results comparison under i.i.d Gaussian noise $\sigma = \{10, 30, 50, 100\}$ on CAVE dataset.}
    \begin{tabular}{ccccccccccccccccc}
    \toprule
    \multirow{2}[4]{*}{Methods} & \multicolumn{4}{c}{$\sigma = 10$} & \multicolumn{4}{c}{$\sigma = 30$} & \multicolumn{4}{c}{$\sigma = 50$} & \multicolumn{4}{c}{$\sigma = 100$} \\
\cmidrule{2-17}          & PSNR  & SSIM  & ERGAS & SAM   & PSNR  & SSIM  & ERGAS & SAM   & PSNR  & SSIM  & ERGAS & SAM   & PSNR  & SSIM  & ERGAS & SAM \\
    \midrule
    Noisy & 28.13  & 0.4371 & 236.40  & 0.7199 & 18.59  & 0.1069 & 676.01  & 1.0085 & 14.15 & 0.0475  & 1126.7 & 1.1461 & 8.13  & 0.0136  & 2253.4 & 1.3074  \\
    \midrule
    PARAFAC & 35.39  & 0.8759 & 108.78  & 0.2363 & 33.65  & 0.8294 & 125.13  & 0.3291 & 31.52 & 0.7393  & 154.19 & 0.4351 & 27.13  & 0.4637  & 250.68 & 0.6681  \\
    \midrule
    LRTA  & 41.36  & 0.9499 & 49.53  & 0.1718 & 36.06  & 0.8775 & 90.50  & 0.2446 & 33.52 & 0.8201  & 121.15 & 0.2897 & 30.06  & 0.7138  & 180.03 & 0.3649  \\
    \midrule
    LRMR  & 39.27  & 0.9094 & 64.81  & 0.3343 & 31.36  & 0.6451 & 157.65  & 0.6021 & 26.67 & 0.4000  & 264.28 & 0.7534 & 26.67  & 0.1850  & 469.26 & 0.9306  \\
    \midrule
    4DHOSVD1 & 45.43  & 0.9811 & 30.83  & 0.1084 & 39.78  & 0.9336 & 59.12  & 0.2272 & 36.82 & 0.8722  & 83.36 & 0.3385 & 32.66  & 0.7307  & 134.34 & 0.5599  \\
    \midrule
    BM4D  & 44.61  & 0.9784 & 33.32  & 0.1289 & 38.80  & 0.9283 & 65.23  & 0.2579 & 35.98  & 0.8685  & 91.19 & 0.3557 & 31.84  & 0.7197  & 144.91 & 0.5160  \\
    \midrule
    TDL   & 44.41  & 0.9797 & 34.32  & 0.1048 & 39.07  & 0.9493 & 63.18  & 0.1493 & 36.46 & 0.9171  & 85.24 & 0.2008 & 32.92  & 0.8284  & 128.15 & 0.3132  \\
    \midrule
    ISTReg & 45.77  & 0.9802 & 30.53  & 0.1086 & 40.51  & 0.9488 & 53.05  & 0.1374 & 37.75 & 0.9271  & 70.16 & 0.1619 & 33.01  & 0.8648  & 120.77 & 0.2376  \\
    \midrule
    LLRT  & 46.60  & 0.9868 & 26.75  & 0.0842 & 41.49  & 0.9681 & 48.50  & 0.1221 & 38.65 & 0.9482  & 67.56 & 0.1551 & 35.39  & 0.9154  & 99.37 & 0.1962  \\
    \midrule
    MSt-SVD & 45.20  & 0.9814 & 32.05  & 0.1064 & 40.23  & 0.9530 & 56.26  & 0.1737 & 37.73 & 0.9285  & 75.03 & 0.2231 & 34.20  & 0.8800  & 115.24 & 0.3142  \\
    \bottomrule
    \end{tabular}%
  \label{table_cave_origin1}%
\end{table*}%

\begin{figure*}[htbp]
\graphicspath{{MSI_image/i.i.d_gaussian/MC_new/}}
\centering
\subfigure[Clean]{
\label{Fig4}
\includegraphics[width=0.88in]{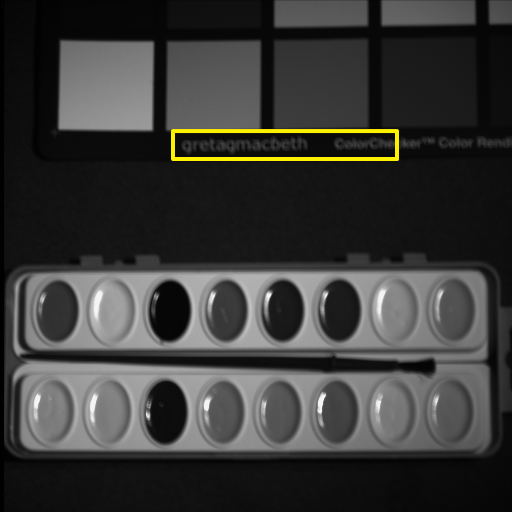}}
\subfigure[Noisy]{
\label{Fig4}
\includegraphics[width=0.88in]{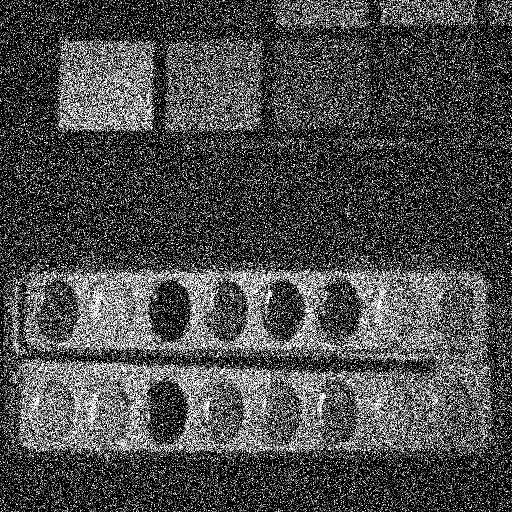}}
\subfigure[4DHOSVD1]{
\label{Fig4}
\includegraphics[width=0.88in]{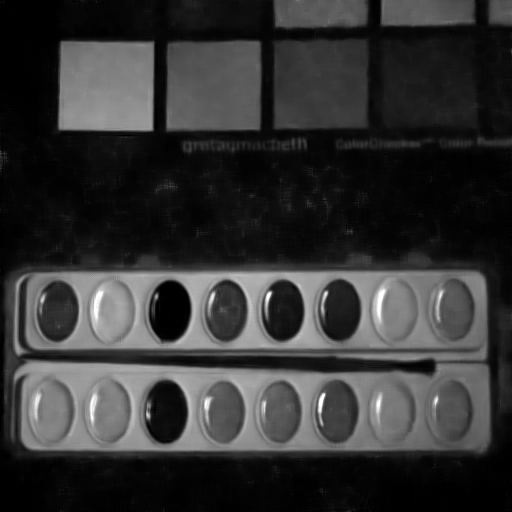}}
\subfigure[TDL]{
\label{Fig4}
\includegraphics[width=0.88in]{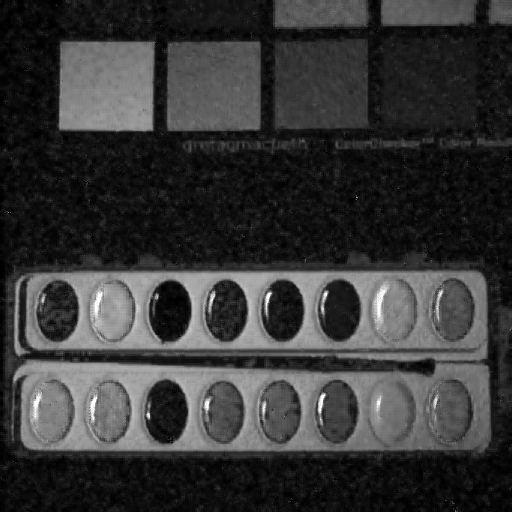}}
\subfigure[BM4D]{
\label{Fig4}
\includegraphics[width=0.88in]{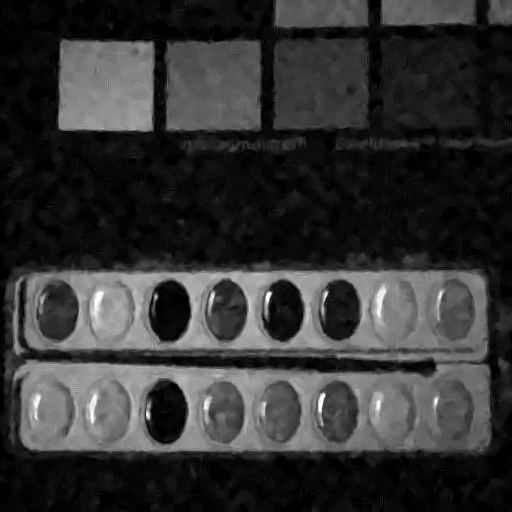}}
\subfigure[LLRT]{
\label{Fig4}
\includegraphics[width=0.88in]{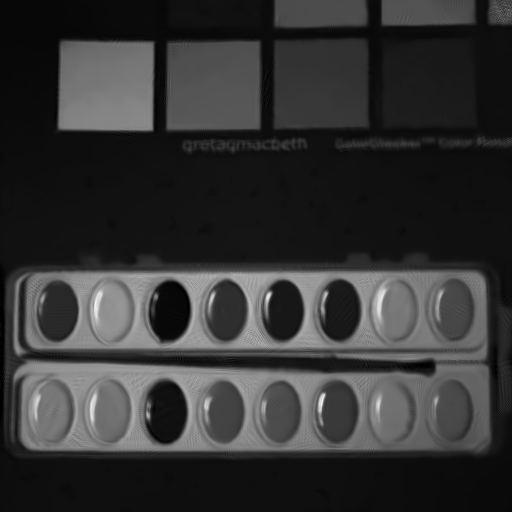}}
\subfigure[MSt-SVD]{
\label{Fig4}
\includegraphics[width=0.88in]{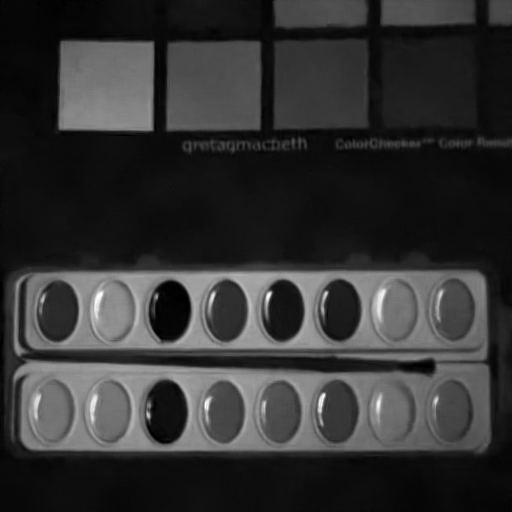}}

\subfigure[Clean]{
\label{Fig4}
\includegraphics[width=1.06in]{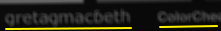}}
\subfigure[4DHOSVD1]{
\label{Fig4}
\includegraphics[width=1.06in]{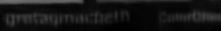}}
\subfigure[TDL]{
\label{Fig4}
\includegraphics[width=1.06in]{TDL_crop}}
\subfigure[BM4D]{
\label{Fig4}
\includegraphics[width=1.06in]{BM4D_crop}}
\subfigure[LLRT]{
\label{Fig4}
\includegraphics[width=1.06in]{LLRT_crop}}
\subfigure[MSt-SVD]{
\label{Fig4}
\includegraphics[width=1.06in]{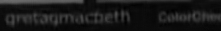}}

\caption{Visual evaluation of five competitive methods on CAVE dataset under i.i.d Gaussian noise with $\sigma = 100$.}

\label{fig_iid_cave}
\end{figure*}

The whole CAVE database consisting of 32 hyperspectral images is used in our synthetic tests. The images of size $512 \times 512 \times 31$ are captured with the wavelengths in the range of 400-700 nm at an interval of 10 nm. In this experiment, entries in all slices were corrupted by zero-mean i.i.d Gaussian noise $N(0,\sigma^2)$ with $\sigma = \{10, 30, 50, 100\}$. Since LRMR and ISTReg require much more memory space, their results are copied from \cite{xu2017multi}. Detailed results are listed in Table \ref{table_cave_origin1}, visual effect comparison is given in Fig. \ref{fig_iid_cave}. It can be seen that the recursive use of patch level information by MSt-SVD may better preserve details.
\subsubsection{non i.i.d. Gaussian Noise}

\begin{table}[htbp]
\ssmall
  \centering
  \caption{Quantitative results comparison under non i.i.d Gaussian noise with input noise level $\sigma = 36$ for all compared methods on CAVE dataset.}
    \begin{tabular}{cccccccc}
    \toprule
    Index & PARAFAC & LRTA & 4DHOSVD1 & BM4D & TDL & LLRT & MSt-SVD \\
    \midrule
    PSNR  & 32.7 & 33.0 & 38.1  & 36.6 & 32.3 & 38.3 & 39.3  \\
    \midrule
    SSIM  & 0.79 & 0.74 & 0.91  & 0.86 & 0.74 & 0.92 & 0.94  \\
    \midrule
    ERGAS & 137.9 & 132.6 & 71.9 & 89.7 & 140.9 & 70.8 & 62.4  \\
    \midrule
    SAM   & 0.40 & 0.45 & 0.33 & 0.40 & 0.48 & 0.21 & 0.19  \\
    \midrule
    Time (m) & 4.8 & 0.6 & 5.5 & 10.1 & 1.2 & 48.8 & 6.2 \\
    \bottomrule
    \end{tabular}%
  \label{table_cave_unbalance}%
\end{table}%

\begin{figure}[htbp]
\graphicspath{{MSI_image/non_i.i.d_gaussian/high_res/}}
\centering
\subfigure[Clean]{
\label{Fig4}
\includegraphics[width=1.018in]{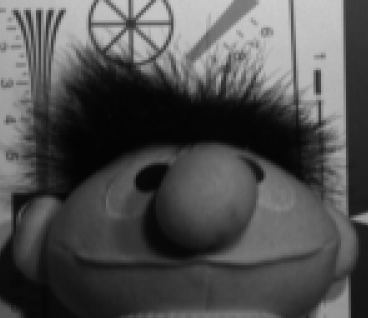}}
\subfigure[4DHOSVD1]{
\label{Fig4}
\includegraphics[width=1.018in]{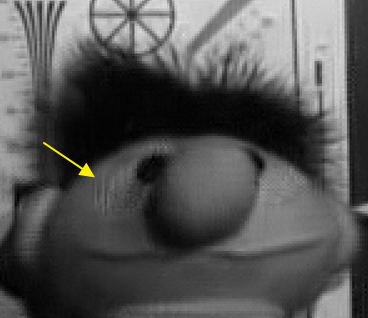}}
\subfigure[TDL]{
\label{Fig4}
\includegraphics[width=1.018in]{TDL_crop}}\\
\subfigure[BM4D]{
\label{Fig4}
\includegraphics[width=1.018in]{BM4D_crop}}
\subfigure[LLRT]{
\label{Fig4}
\includegraphics[width=1.018in]{LLRT_crop}}
\subfigure[MSt-SVD]{
\label{Fig4}
\includegraphics[width=1.018in]{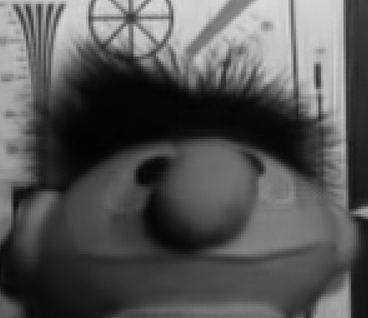}}

\caption{Visual evaluation of compared methods under non i.i.d Gaussian noise.}

\label{fig_non_iid_cave}
\end{figure}

Entries in all slices were corrupted by zero-mean Gaussian noise with increasing intensity from 21 to 51, and for fair comparison, the input noise level for all compared methods is taken as the average number 36. The detailed results and visual evaluations are given in Table \ref{table_cave_unbalance}\footnote{BM4D uses a C++ mex-function with parallel implementation and TDL uses some mature toolboxes.} and Fig. \ref{fig_non_iid_cave}, respectively. Comparing Table \ref{Table_cc15} and Table \ref{table_cave_unbalance}, it is interesting to see that compared with 4DHOSVD, MSt-SVD is faster in dealing with color images but slower in multispectral images, this is because the for-loop of Matlab slice-by-slice computation in the Fourier domain is slow, and our C++ mex-function can reduce the total time to 1.5 minutes.
\subsubsection{Experiments on HHD data}
\begin{figure}[htbp]
\graphicspath{{MSI_image/Real_HDD/Real_HDD_new/}}
\centering
\subfigure[Noisy]{
\label{Fig4}
\includegraphics[width=1.068in]{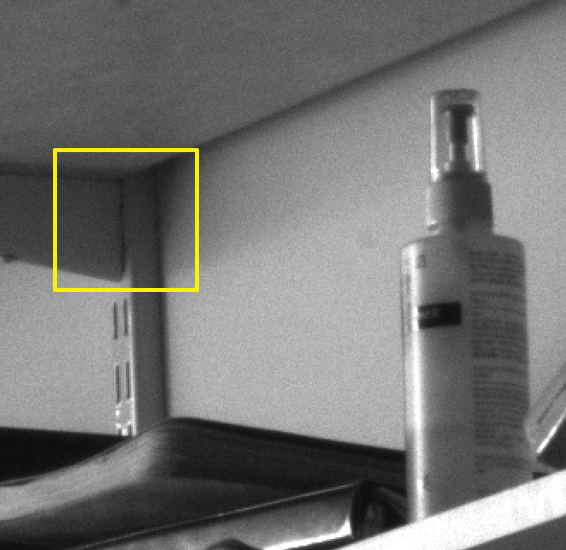}}
\subfigure[TDL]{
\label{Fig4}
\includegraphics[width=1.068in]{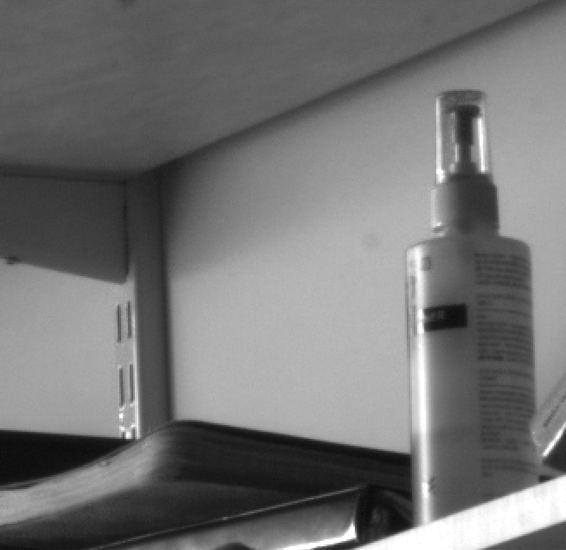}}
\subfigure[MSt-SVD]{
\label{Fig4}
\includegraphics[width=1.068in]{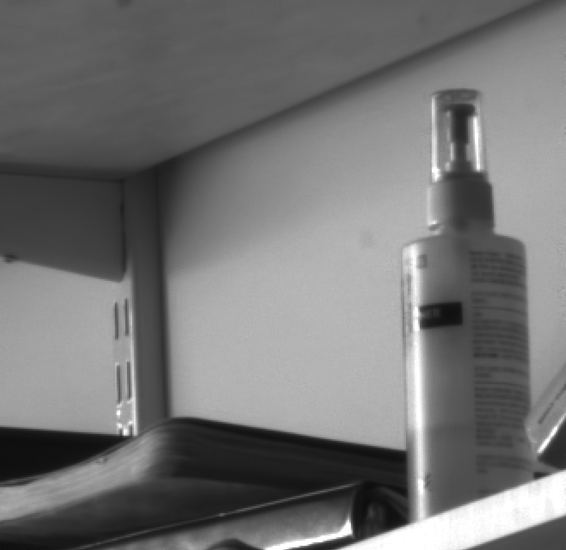}}\\
\subfigure[Noisy]{
\label{Fig4}
\includegraphics[width=1.068in]{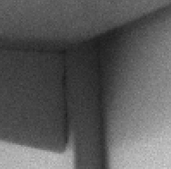}}
\subfigure[TDL]{
\label{TDL_imgd3_crop_paint}
\includegraphics[width=1.068in]{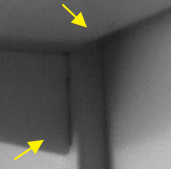}}
\subfigure[MSt-SVD]{
\label{Fig4}
\includegraphics[width=1.068in]{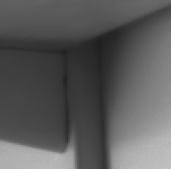}}\\
\subfigure[Noisy]{
\label{Fig4}
\includegraphics[width=1.068in]{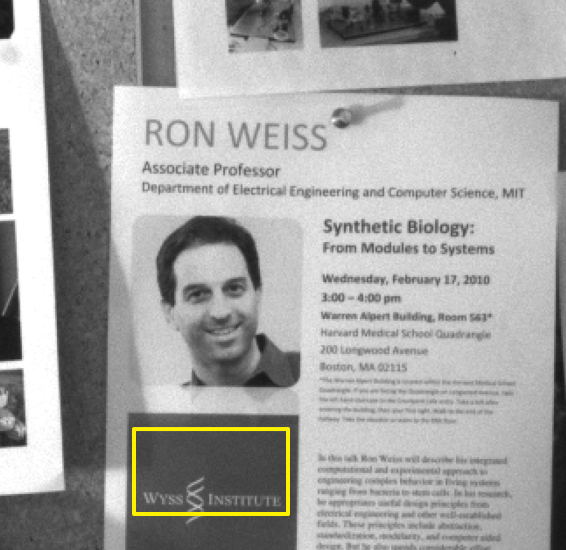}}
\subfigure[TDL]{
\label{Fig4}
\includegraphics[width=1.068in]{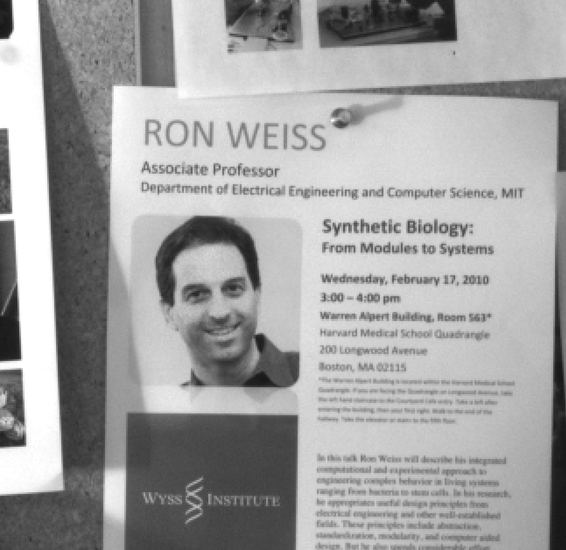}}
\subfigure[MSt-SVD]{
\label{Fig4}
\includegraphics[width=1.068in]{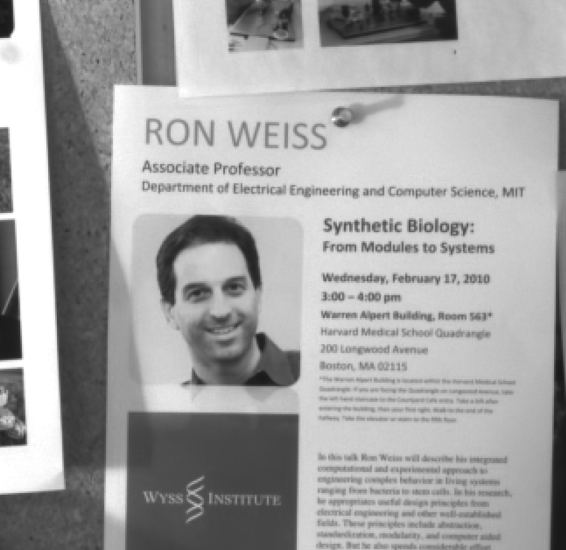}}\\
\subfigure[Noisy]{
\label{Fig4}
\includegraphics[width=1.068in]{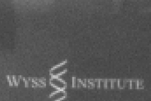}}
\subfigure[TDL]{
\label{TDL_imgd7_crop}
\includegraphics[width=1.068in]{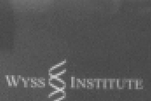}}
\subfigure[MSt-SVD]{
\label{Fig4}
\includegraphics[width=1.068in]{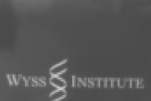}}

\caption{Visual evaluation of TDL and MSt-SVD on real HHD dataset. Please zoom in for better view.}

\label{fig_real_HDD}
\end{figure}

There are 50 images of size $1040 \times 1392 \times 31$, and some of them are clearly contaminated by noise. Considering the large size of noisy images, we mainly examine the effectiveness of MSt-SVD on handling real multispectral image, and compare it with the efficient benchmark TDL. The input noise level $\sigma$ is manually tuned for both TDL and MSt-SVD to balance smooth effects and details, and $N_{step} = 8$ is used for MSt-SVD to save some time. Visual evaluation is given in Fig. \ref{fig_real_HDD}, and the artifacts produced by TDL can be seen in Fig. \ref{TDL_imgd3_crop_paint} and Fig. \ref{TDL_imgd7_crop}.
\subsubsection{Experiments on Urban HSI data}
\begin{figure}[htbp]
\graphicspath{{MSI_image/Real_urban/high_res/}}
\centering
\subfigure[Noisy]{
\label{Fig4}
\includegraphics[width=0.78in]{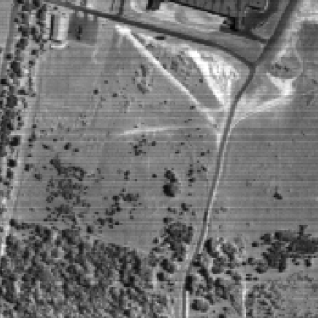}}
\subfigure[MSt-SVD]{
\label{Fig4}
\includegraphics[width=0.78in]{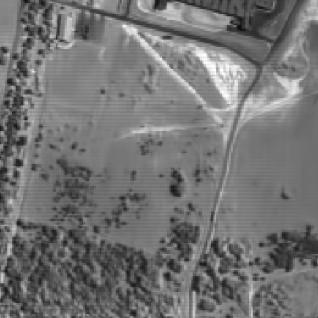}}
\subfigure[Nmog]{
\label{Fig4}
\includegraphics[width=0.78in]{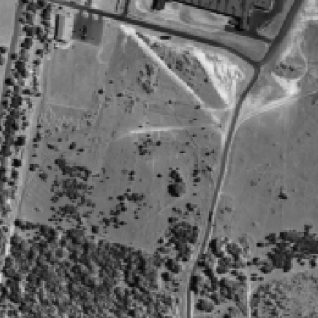}}
\subfigure[twist MSt-SVD]{
\label{Fig4}
\includegraphics[width=0.78in]{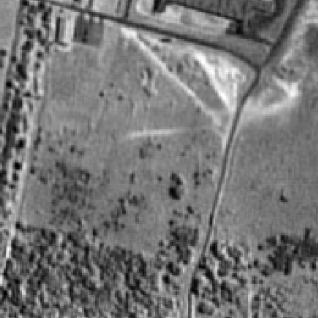}}

\caption{Visual evaluation on band 103 in Urban data.}

\label{fig_real_urban1}
\end{figure}

\begin{figure}[htbp]
\graphicspath{{MSI_image/Real_urban/high_res/}}
\centering
\subfigure[Noisy]{
\label{Fig4}
\includegraphics[width=1.018in]{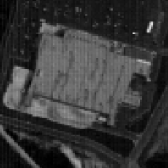}}
\subfigure[Nmog]{
\label{Fig4}
\includegraphics[width=1.018in]{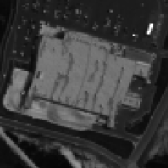}}
\subfigure[twist MSt-SVD]{
\label{Fig4}
\includegraphics[width=1.018in]{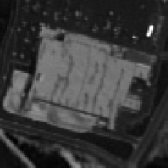}}

\caption{Visual evaluation on band 1 in Urban data.}

\label{fig_real_urban2}
\end{figure}

The full Urban dataset of size $307\times 307 \times 210$ is used, and some of the bands are contaminated by stripe noise in Fig. \ref{fig_real_urban1}(a) and Fig. \ref{fig_real_urban2}(a). As illustrated in \ref{fig_real_urban1}(b), naively applying MSt-SVD fails to remove stripe noise. In fact, the influence of the sparse stripe noise may be amplified by recursively computing the row- and column-wise relationship. According to (\ref{group_equal}), the use of block representation ($\ref{bcirc_matrix}$) does not change the group level representation, thus the influence of stripe noise could be attenuated by adopting sparsity in the grouping dimension. Specifically, we first reshape the original data as a new image of size $210 \times 307 \times 307$, such that stripe noise sparsely spread along the grouping dimension, and then apply MSt-SVD to the new data, the final filtered image is obtained by reshaping it back to the original size. We term this simple operation 'twist MSt-SVD' and notice that it does not increase computational burden. Since many compared methods can not handle the stripe noise \cite{chen2018denoising}, we use the benchmark Nmog for visual effects comparison. Fig. \ref{fig_real_urban1}(d) and Fig. \ref{fig_real_urban2}(c) demonstrate the effectiveness of the twist implementation. Considering its efficiency, its performance may be improved by further modeling of tensor sparsity \cite{qi2018multi}.
\subsection{Choice of parameters}
\begin{figure}[htbp]
\graphicspath{{Discussion/}}

  \centering
  \includegraphics[width=2.6in]{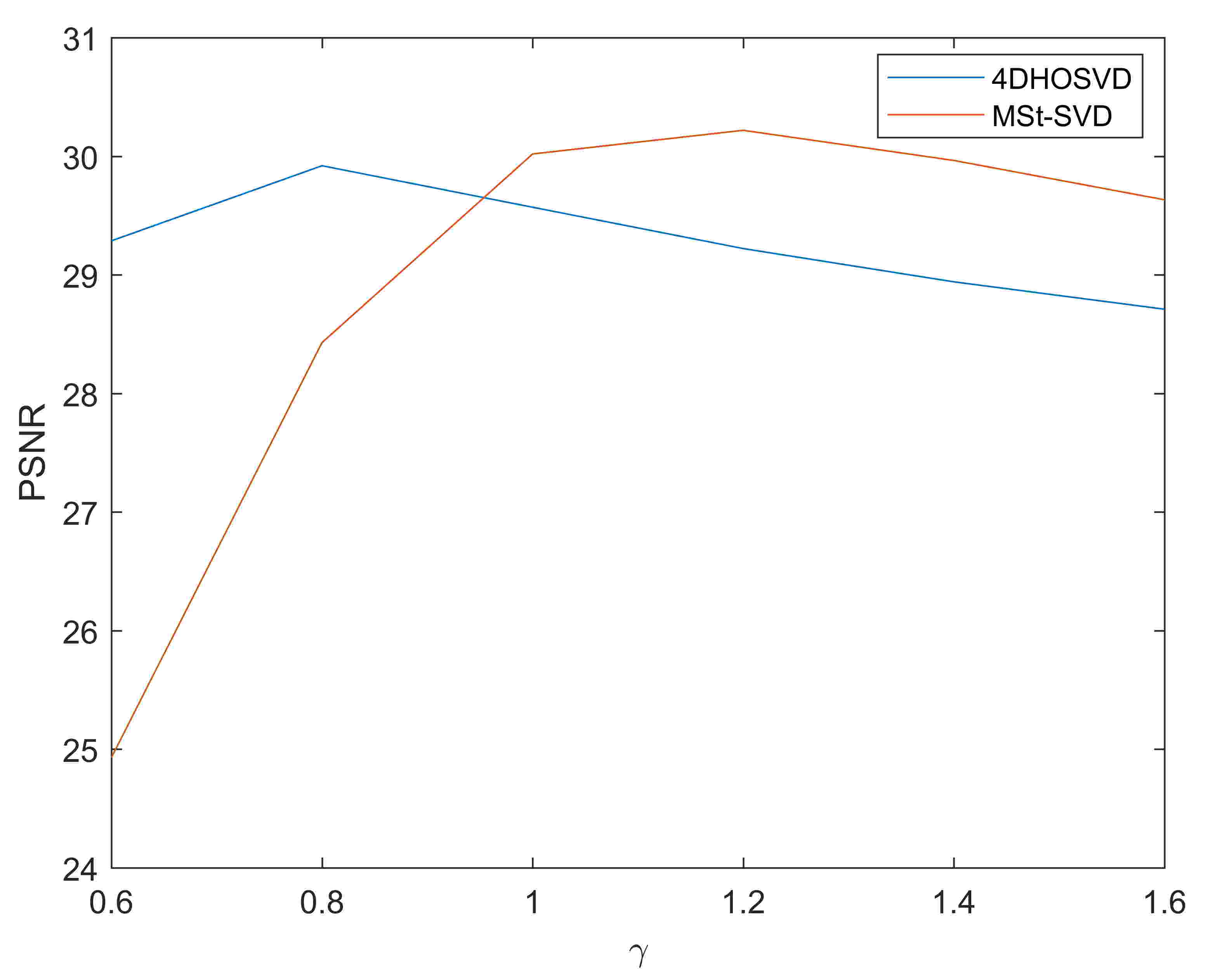}

\caption{Average PSNR values of 4DHOSVD and MSt-SVD on the first 8 images of Kodak gallery when $\sigma = 30$ under different hard-threshold scale factor $\gamma$.}

\label{fig_tau_tune}
\end{figure}

\begin{figure}[htbp]
\graphicspath{{Discussion/}}
\centering
\subfigure[Clean]{
\label{Fig4}
\includegraphics[width=0.68in]{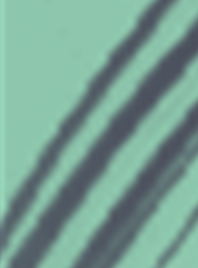}}
\subfigure[Noisy]{
\label{Fig4}
\includegraphics[width=0.68in]{noisy_crop}}
\subfigure[$\gamma = 1.2$]{
\label{Fig4}
\includegraphics[width=0.68in]{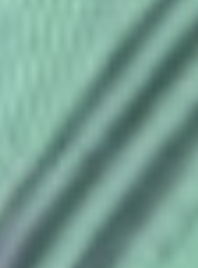}}
\subfigure[$\gamma = 1.4$]{
\label{Fig4}
\includegraphics[width=0.68in]{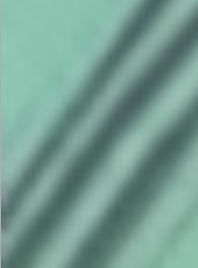}}

\caption{The influence of hard-threshold parameter on MSt-SVD when $\sigma = 60$.}

\label{fig_tau_tune_zoom_in}
\end{figure}

Among all free parameters, the hard-threshold parameter $\tau$ directly controls the core tensor sparsity in the transform domain, so we mainly investigate how $\tau$ could influence the proposed MSt-SVD and choose $ps = 8$, $SR = 20$ and $K$ = 30 according to the settings of 4DHOSVD in \cite{rajwade2013image}. For 4DHOSVD in \cite{rajwade2013image}, the authors set $\tau = \sigma\sqrt{2log(n\_elem)}$ based on \cite{donoho1994ideal} for their simulated experiments on Kodak gallery\footnote{http://r0k.us/graphics/kodak/}, where $n\_elem$ is the number of elements of a patch group. But we observe that it is chosen too large to provide over-smooth effects, so we multiply $\tau$ with a scale factor $\gamma$, and tune the averagely best PSNR value using the first 8 images of Kodak gallery. Fig. \ref{fig_tau_tune} shows the influence of $\gamma$ on both 4DHOSVD and MSt-SVD, and the choice of parameters is detailed in Table \ref{table_para_choice}. Fig. \ref{fig_tau_tune_zoom_in} gives an illustration of the influence of $\tau$ with Brodatz color texture\footnote{http://multibandtexture.recherche.usherbrooke.ca/colored\_brodatz\_more.html}. In real cases, the tuning of $\tau$ for MSt-SVD is efficient because some intermediate results can be preserved to avoid the recursive computation of grouping and local PCA transform.

\begin{table}[htbp]
\ssmall
  \centering
  \caption{Choice of parameters for 4DHOSVD and MSt-SVD on color image and multispectral image (MSI) experiments.}
    \begin{tabular}{ccccc}
    \toprule
    \multirow{2}[4]{*}{Parameter} & \multicolumn{2}{c}{4DHOSVD} & \multicolumn{2}{c}{MSt-SVD} \\
\cmidrule{2-5}          & Color Image & MSI   & Color Image & MSI \\
    \midrule
    $ps$    & 8    & 8    & 8     & 8  \\
    \midrule
    $K$     & 30    & 30    & 30    & 30 \\
    \midrule
    $SR$    & 20    & 16    & 20    & 16 \\
    \midrule
    $N_{step}$ & 4     & 4     & 4     & 4 \\
    \midrule
    $\gamma$ & 0.8     & 1     & 1.1 ($\sigma<30$), 1.2 ($\sigma>30$)    & 4 \\
    \bottomrule
    \end{tabular}%
  \label{table_para_choice}%
\end{table}%

\section{Conclusion}
In this paper, we build the relationship among state-of-the-art transforms with block diagonal representation, and investigate the proper choice of patch level transform. According to our discussion and analysis, two simple and effective methods that combine a global t-SVD basis and local PCA transform are proposed. The proposed MSt-SVD and CMSt-SVD utilize more spatial information, and produce competitive performance with state-of-the-art filters in terms of both efficiency and effectiveness. Recently, some statistical properties of tensor decomposition \cite{Zhang2018Tensor} are studied, it is interesting to investigate a further understanding of both color image and multispectral image denoising with block diagonal representation. Besides, our future research also includes classification \cite{Lu2008MPCA} and related image restoration problems \cite{tang2017pairwise}.


\begin{appendices}
\section{Some Features Related To Block Circulant Representation (\ref{bcirc_matrix})}
\label{appendix_a}
\begin{myThm}
$bcirc(\mathcal{P}_i) bcirc(\mathcal{P}_i)^T$ is also a block circulant matrix.
\label{circ_times_transpose}
\end{myThm}

\begin{myThm}
Two patches $\mathcal{P}_i$ and $\mathcal{P}_j$ are similar if and only if $bcirc(\mathcal{P}_i)$ and $bcirc(\mathcal{P}_j)$ are similar. More specifically,
\begin{equation}\label{norm_equal}
  \|\mathcal{P}_i - \mathcal{P}_j\|_F = \sqrt{3}\|bcirc(\mathcal{P}_i) - bcirc(\mathcal{P}_j)\|_F
\end{equation}
\label{circ_norm}
\end{myThm}

\begin{myThm}
Given a group of similar patches $\mathcal{G}$, and its block circulant tensor representation $bcirc(\mathcal{G})$ in (\ref{bcirc_tensor}). If $\mathbf{U}_{group}$ and $\mathbf{U}_{bcirc\_group}$ are the last mode factor matrix of $\mathcal{G}$ and $bcirc(\mathcal{G})$, respectively, then
\begin{equation}\label{group_equal}
  \mathbf{U}_{group} = \mathbf{U}_{bcirc\_group}
\end{equation}
\label{circ_group_equal}
\end{myThm}
\noindent The proof of above Theorem is not hard by checking corresponding definition. Theorem \ref{circ_times_transpose} offers an efficient implementation to compute product between a block circulant matrix and its transpose. Theorem \ref{circ_norm} and Theorem \ref{circ_group_equal} indicate that nonlocal similarity and linear representation do not change after block circulant operation.

\section{Re-formulation of Block Circulant Tensor Decomposition In The Fourier Domain}
\label{appendix_b}
\begin{myThm}
\label{theo_bcirc_diag}
\cite{rojo2004some} Given a patch $\mathcal{P}_i$ and its block circulant representation $bcirc(\mathcal{P})$ in ($\ref{bcirc_matrix}$), there exists an orthogonal matrix $\mathbf{F}\otimes\mathbf{I}$ that could transform $bcirc(\mathcal{P})$ into a block diagonal matrix via
\begin{equation}\label{bcirc_diag_trans}
  bdiag(\hat{\mathcal{P}}) = (\mathbf{F}\otimes\mathbf{I})bcirc(\mathcal{P})(\mathbf{F}\otimes\mathbf{I})^{-1}
\end{equation}
where $\otimes$ represents the kronecker product, and $\hat{\mathcal{P}} = \mathcal{P} \times _3\mathbf{W}$, $\mathbf{W} = \sqrt{3} \mathbf{F}$. Furthermore, according to the orthogonality, we have
\begin{equation}\label{core_tensor_equal}
  \|\mathcal{C}_{fdiag}\|_F =  \|\mathcal{C}_{bcirc}\|_F
\end{equation}
\end{myThm}

\end{appendices}

\section*{Acknowledgment}

The authors would like to thank all the reviewers for their valuable comments, and appreciate all the authors for sharing their code and software package.



%
\bibliographystyle{IEEEtran}
\bibliography{IEEEabrv,ms}


\end{document}